\documentclass[11pt, a4paper, logo, copyright]{googlecloud}

\pdftrailerid{redacted}

\makeatletter
\renewcommand\bibentry[1]{\nocitep{#1}{\frenchspacing\@nameuse{BR@r@#1\@extra@b@citeb}}}
\makeatother

\usepackage{kantlipsum, lipsum}
\usepackage{dsfont}
\usepackage[authoryear, sort&compress, round]{natbib}

\usepackage[utf8]{inputenc} %
\usepackage[T1]{fontenc}    %
\usepackage{hyperref}       %
\hypersetup{
  colorlinks,
  citecolor=blue,
  linkcolor=red,
  urlcolor=blue}
\usepackage{booktabs}       %
\usepackage{nicefrac}       %
\usepackage{microtype}      %
\usepackage{wrapfig} %
\usepackage{soul} %
\usepackage{tikz, adjustbox}
\usepackage{arydshln}
\usepackage[capitalize,noabbrev]{cleveref}
\usepackage{fancyvrb}

\usepackage{inconsolata}
\usepackage[skins,breakable,most]{tcolorbox}

\usepackage{caption}
\usepackage{subcaption}
\usepackage{enumitem}
\usepackage{flushend}
\usepackage{balance}
\usepackage{lineno}
\usepackage{amsmath,amssymb,amsfonts}
\usepackage{algorithm}
\usepackage{algpseudocode}
\usepackage{graphicx}
\usepackage{textcomp}
\usepackage{xcolor}
\usepackage{xspace}
\usepackage{colortbl}
\usepackage{amsthm}
\usepackage{url}
\usepackage{float}
\usepackage{multirow}
\usepackage{multicol}
\usepackage{color}
\usepackage{bm}
\usepackage{bbm}

\usepackage{pifont}

\usepackage{array}
\newcolumntype{L}[1]{>{\raggedright\let\newline\\\arraybackslash\hspace{0pt}}m{#1}}
\newcolumntype{C}[1]{>{\centering\let\newline  \\\arraybackslash\hspace{0pt}}m{#1}}
\newcolumntype{R}[1]{>{\raggedleft\let\newline \\\arraybackslash\hspace{0pt}}m{#1}}

\usepackage{titletoc}
\newcommand\DoToC{%
  \startcontents
    \printcontents{}{1}{\textbf{Contents of Appendix}\vskip3pt\hrule\vskip5pt}
  \vskip3pt\hrule\vskip5pt
}
\newcommand{\ours}{\textsc{SkillOS}\xspace}
\newcommand{\skillos}{
{\bfseries
\textcolor[HTML]{4285F4}{S}%
\textcolor[HTML]{EA4335}{k}%
\textcolor[HTML]{FBBC05}{i}%
\textcolor[HTML]{4285F4}{l}%
\textcolor[HTML]{34A853}{l}%
\textcolor[HTML]{FBBC05}{O}%
\textcolor[HTML]{34A853}{S}}\xspace
}

\usepackage{xcolor}

\newtcbox{\oplabel}[2][]{
  on line,
  boxsep=1pt,
  left=3pt,
  right=3pt,
  top=1pt,
  bottom=1pt,
  arc=1pt,
  boxrule=0pt,
  colback=#2!12,
  coltext=#2!85!black,
  fontupper=\bfseries\scriptsize,
  #1
}

\definecolor{insertpink}{HTML}{D9308A}
\definecolor{updategreen}{HTML}{188038}
\definecolor{deleteorange}{HTML}{F29900}

\newcommand{\insertop}{\oplabel{insertpink}{insert\_skill}}
\newcommand{\updateop}{\oplabel{updategreen}{update\_skill}}
\newcommand{\deleteop}{\oplabel{deleteorange}{delete\_skill}}

\definecolor{lightorange}{RGB}{245, 237, 211}
\definecolor{clovergreen}{RGB}{32,115,55}
\definecolor{redlinkcolor}{rgb}{0.79607843, 0.25098039, 0.25882353}
\definecolor{bluecitecolor}{rgb}{0,0.36,0.69}
\definecolor{lightorange}{RGB}{245, 237, 211} 
\definecolor{bluebar}{RGB}{138,159,201}
\definecolor{pinkbar}{RGB}{232,180,189}
\definecolor{deepgreen}{RGB}{152,206,187}
\definecolor{lightgreen}{RGB}{224,238,190}
\definecolor{googleblue}{RGB}{87,134,236}
\definecolor{stdgray}{gray}{0.5}
\newcommand{\valstd}[2]{$#1_{\,\textcolor{stdgray}{\scriptscriptstyle #2}}$}

\newtcbox{\hlprimarytab}{on line, rounded corners, box align=base, colback=c3!10,colframe=white,size=fbox,arc=3pt, before upper=\strut, top=-2pt, bottom=-4pt, left=-2pt, right=-2pt, boxrule=0pt}
\newtcbox{\hlsecondarytab}{on line, box align=base, colback=blue!10,colframe=white,size=fbox,arc=3pt, before upper=\strut, top=-2pt, bottom=-4pt, left=-2pt, right=-2pt, boxrule=0pt}
\newtcbox{\hlcasetab}{on line, box align=base, colback=c5!10,colframe=white,size=fbox,arc=3pt, before upper=\strut, top=-2pt, bottom=-4pt, left=-2pt, right=-2pt, boxrule=0pt}

\definecolor{c1}{cmyk}{0,0.6175,0.8848,0.1490} 
\definecolor{c2}{cmyk}{0.1127,0.6690,0,0.4431} 
\definecolor{c3}{cmyk}{0.3081,0,0.7209,0.3255} 
\definecolor{c4}{cmyk}{0.6765,0.2017,0,0.0667} 
\definecolor{c5}{cmyk}{0,0.8765,0.7099,0.3647}

\usepackage[utf8]{inputenc} 
\usepackage[T1]{fontenc}    
\usepackage{amsfonts}       
\usepackage{comment} 
\usepackage{mdframed}
\usepackage{fontawesome}
\usepackage{csquotes}
\usepackage{makecell}
\definecolor{beigecolor}{RGB}{253, 244, 204} 
\definecolor{greencolor}{RGB}{228, 242, 217} 
\definecolor{bluecolor}{RGB}{66, 133, 244} 
\definecolor{orgcolor}{RGB}{255, 140, 15} 

\definecolor{redcolor}{RGB}{234, 67, 53} 

\definecolor{ggreen}{RGB}{52, 168, 83}

\definecolor{gyellow}{RGB}{251, 188, 5}

\definecolor{lightorange}{RGB}{245, 237, 211} 
\definecolor{bluebar}{RGB}{138,159,201}
\definecolor{pinkbar}{RGB}{232,180,189}

\usepackage{mathtools}

\lstdefinestyle{mystyle}{
    backgroundcolor=\color{backcolour},   
    commentstyle=\color{codegreen},
    keywordstyle=\color{magenta},
    numberstyle=\tiny\color{codegray},
    stringstyle=\color{codepurple},
    basicstyle=\ttfamily\scriptsize,
    breakatwhitespace=false,         
    breaklines=true,                 
    captionpos=b,                    
    keepspaces=true,                 
    numbers=left,                    
    numbersep=5pt,                  
    showspaces=false,                
    showstringspaces=false,
    showtabs=false,                  
    tabsize=2,
    frame=none,
    aboveskip=1pt,
    belowskip=1pt,
}
\lstdefinestyle{plainins}{
    backgroundcolor=\color{white},   
    commentstyle=\color{codegreen},
    keywordstyle=\color{magenta},
    numberstyle=\tiny\color{codegray},
    stringstyle=\color{codepurple},
    basicstyle=\ttfamily\scriptsize,
    breakatwhitespace=false,         
    breaklines=true,                 
    captionpos=b,                    
    keepspaces=true,                 
    numbers=none,                    
    numbersep=5pt,                  
    showspaces=false,                
    showstringspaces=false,
    showtabs=false,                  
    tabsize=2,
    aboveskip=0pt,
    belowskip=0pt,
    frame=single
}
\lstdefinestyle{plainexam}{
    backgroundcolor=\color[HTML]{FFFCF3},   
    commentstyle=\color{codegreen},
    keywordstyle=\color{magenta},
    numberstyle=\tiny\color{codegray},
    stringstyle=\color{codepurple},
    basicstyle=\ttfamily\scriptsize,
    breakatwhitespace=false,         
    breaklines=true,                 
    captionpos=b,                    
    keepspaces=true,                 
    numbers=none,                    
    numbersep=5pt,                  
    showspaces=false,                
    showstringspaces=false,
    showtabs=false,                  
    tabsize=2,
    aboveskip=0pt,
    belowskip=0pt
}

\title{\skillos{}: Learning Skill Curation for Self-Evolving Agents}

\correspondingauthor{siruo2@illinois.edu, \{junyann, chenyulee\}@google.com}

\renewcommand{\today}{}

\author[1*]{Siru Ouyang}
\author[2$\dagger$]{Jun Yan}
\author[2]{Yanfei Chen}
\author[2]{Rujun Han}
\author[2]{Zifeng Wang}
\author[2]{Bhavana Dalvi Mishra}
\author[2]{Rui Meng}
\author[2]{Chun-Liang Li}
\author[1]{Yizhu Jiao}
\author[3]{Kaiwen Zha}
\author[3]{Maohao Shen}
\author[2]{Vishy Tirumalashetty}
\author[2]{George Lee}
\author[1]{Jiawei Han}
\author[2]{Tomas Pfister}
\author[2$\dagger$]{Chen-Yu Lee}

\affil[1]{University of Illinois Urbana-Champaign}
\affil[2]{Google Cloud AI Research}
\affil[3]{Massachusetts Institute of Technology}

\begin{abstract}
LLM-based agents are increasingly deployed to handle streaming tasks, yet they often remain one-off problem solvers that fail to learn from past interactions. Reusable skills distilled from experience provide a natural substrate for self-evolution, where high-quality skill curation serves as the key bottleneck. Existing approaches either rely on manual skill curation, prescribe heuristic skill operations, or train for short-horizon skill adaptation, but still struggle to learn complex long-term curation policies from indirect and delayed feedback.
We propose\skillos, an experience-driven RL training recipe for learning skill curation in self-evolving agents. \ours{} pairs a frozen \textit{agent executor} that retrieves and applies skills with a trainable \textit{skill curator} that updates an external \textsc{SkillRepo} from accumulated experience. To provide learning signals for curation, we train on grouped task streams based on skill-relevant task dependencies, where earlier trajectories update the \textsc{SkillRepo}, and later related tasks evaluate these updates. We further design composite rewards to better attribute downstream executor feedback to curation decisions.
Across multi-turn agentic tasks and single-turn reasoning tasks, \ours{} consistently outperforms memory-free and strong memory-based baselines in both effectiveness and efficiency, with the learned skill curator generalizing across different executor backbones and task domains. Further analyses show that the learned curator produces more targeted skill use, while the evolving \textsc{SkillRepo} develops richer internal structure and higher-level meta-skills over time.
\end{abstract}

\begin{document}

\maketitle

\section{Introduction}

LLM-based agents~\citep{DBLP:journals/fcsc/WangMFZYZCTCLZWW24} are increasingly deployed in real-world scenarios, where they must move beyond instantaneous problem-solving toward long-term proficiency~\citep{he2026memoryarena}. However, the prevailing paradigm of ``one-off'' task execution limits their utility in streaming settings, where tasks unfold sequentially over time. This makes \emph{self-evolution}~\citep{fang2025comprehensive, gao2025survey} essential: capable agents should not repeatedly start from scratch, but instead continually accumulate, refine, and reuse experience for future tasks.

A key substrate for self-evolution is \emph{procedural memory}~\citep{hu2025memory, wu2025human, DBLP:journals/corr/abs-2508-06433}, specifically, reusable skills~\citep{anthropic_skills_2025, wang2025inducing} accumulated from past interactions. In real-world streaming settings~\citep{wu2024streambench}, a skill-based self-evolving agent typically follows a closed-loop workflow: for each new task, it selects relevant skills, uses them to guide execution, and updates its skill collection based on the resulting trajectory. This makes skill curation---the extraction of high-quality lessons and their integration into the skill collection---essential for self-evolving agents.

However, existing skill curation works remain limited. Manually curated skills, such as Anthropic's skills repository~\citep{anthropic_skills_2025}, demand huge human expertise and cannot scale to the diversity of tasks that agents may encounter. Prompting or heuristic-based methods that dictate memory operations~\citep{xu2025amem, qiu2025alita, DBLP:journals/corr/abs-2504-07079} rely on fixed rules and lack downstream performance feedback, preventing them from adapting to the executor's actual needs. Recent studies explored reinforcement learning (RL) to optimize skill-based agent systems. However, they either focus on teaching agents to \textit{use} skills~\citep{xia2026skillrl, tu2026dynamic} or
optimize skill operations within a short task stream~\citep{DBLP:journals/corr/abs-2512-17102, DBLP:journals/corr/abs-2602-10652}. This limits the density of learning signals available for curating highly reusable skills and mastering complex management operations such as skill update and deletion, which are essential for robust and scalable long-term self-evolution.

\begin{figure}[t]
\begin{center}
\includegraphics[width=0.98\textwidth]{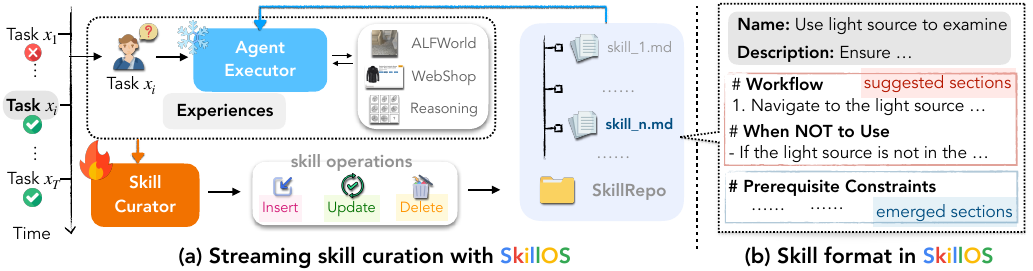}
\end{center}
\vspace{-3mm}
\caption{\ours{} pairs a frozen \textit{Agent Executor} with a trainable \textit{Skill Curator}. The executor retrieves relevant skills from \textsc{SkillRepo} to act; the curator edits the repo (insert/update/delete) based on the resulting experiences, with Markdown as the skill format.
}
\label{fig: skillos_overview}
\vspace{-5mm}
\end{figure}

To tackle this challenge, we propose\skillos, an experience-driven RL training recipe to learn the capability of skill curation for self-evolving agents. 
We study skill curation in a modular multi-agent framework in a streaming setting, where a frozen \emph{agent executor} solves tasks with a skill collection (termed \textsc{SkillRepo}), while a trainable \emph{skill curator} updates and manages this collection through function calls (Figure~\ref{fig: skillos_overview}\textcolor{redlinkcolor}{(a)}).
We represent skills as Markdown files~\citep{anthropic_skills_2025} (Figure~\ref{fig: skillos_overview}\textcolor{redlinkcolor}{(b)}) managed via file I/O operations similar to an operating system (OS).
Our recipe features two core designs. 
\textit{First}, we construct each training instance as a group of related tasks. By mimicking test-time streaming settings, it grounds skill curation in long-term utility: skills induced from earlier experiences are evaluated by their ability to improve later related tasks.
\textit{Second}, we design rewards to better attribute environmental feedback to curation decisions, combining task performance with signals for valid function calls, skill quality, and \textsc{SkillRepo}'s compactness. Together, these designs turn delayed and indirect supervision into learning signals for skill curation.

We evaluate \ours{} on both multi-turn agentic tasks and single-turn reasoning tasks. Experiments show that \ours{} consistently outperforms memory-free and strong memory-based methods in both effectiveness and efficiency, with up to $+9.8\%$ relative performance improvement and $-6.0\%$ fewer interaction steps compared to the strongest baseline (Table~\ref{table: alfworld}). Our trained skill curator generalizes well across executors and tasks, improving performance even with the Gemini-2.5-Pro executor. Notably, our 8B curator also outperforms Gemini-2.5-Pro when used directly as the curator. Beyond performance gains, our analyses further show that the learned skill curator leads to more targeted and effective skill utilization, while the skills in \textsc{SkillRepo} evolve into more richly structured Markdown files that encode higher-level meta-skills over time. Together, we establish \ours{} as a practical, modular, and experience-driven RL training recipe for building self-evolving agents.
\section{Related Work}

\noindent\textbf{Memory for Self-Evolving Agents.} 
Learning from past experiences as procedural memory~\citep{wu2025human, wei2025evo, shen2026decocted, hu2025memory, huang2026rethinking, zhang2024working} is a central mechanism for developing self-evolving agents~\citep{gao2025survey, fang2025comprehensive}. 
The central challenge is to encode interaction histories into reusable and retrievable representations.
Case-based representations are the most concrete form in this research line: they store experiences in minimally processed formats, allowing past histories to be replayed directly or reused as in-context exemplars, such as raw trajectories~\citep{zheng2023synapse, DBLP:journals/corr/abs-2508-16153, wu2025comemagent} and abstracted query–response pairs~\citep{zhao2024expel, islam-etal-2024-mapcoder}. Another line of work abstracts experiences into higher-level knowledge that is editable, auditable, and composable, reducing reliance on long trajectory replay and improving both cross-task generalization and efficiency. Such strategy-based memory typically consists of reusable workflows~\citep{wang2025agent, DBLP:journals/corr/abs-2507-06229}, distilled insights~\citep{ouyang2026reasoningbank, huang-etal-2025-r2d2, DBLP:journals/corr/abs-2509-04439}, and recurring patterns~\citep{yang2024buffer, kim-etal-2025-principles}.
Recently, skills~\citep{wang2025inducing, kuroki2025agent, DBLP:journals/corr/abs-2602-08004, DBLP:journals/corr/abs-2602-12670, DBLP:journals/corr/abs-2602-02474, yang2026autoskillexperiencedrivenlifelonglearning, alzubi2026evoskill, liang2026skillnet} have emerged as a new agent-native form of memory and an orchestrable capability layer, owing to their modularity and ease of customization. Anthropic conceptualizes each skill as a folder containing instructions, scripts, and supporting resources~\citep{anthropic_agent_skills_overview}, which has become the most widely adopted design in the current community. Our work follows this design philosophy, simplifying the setting for research purposes by representing each skill as a single \textit{Markdown} file.

\noindent\textbf{Learning Memory and Skill Curation with RL.} 
Training LLM-based agent systems with memory capabilities using RL has become a growing research direction. One research line targets training for long-context management with predefined operations such as compaction~\citep{zhou2026mem, yu2026memagent, wang2025mem}. Another interesting area focuses more on memory utilization and management by learning additional memory tool-calls~\citep{DBLP:journals/corr/abs-2508-19828, DBLP:journals/corr/abs-2508-16629, DBLP:journals/corr/abs-2510-12635} or training policies for different stages, such as memory retrieval~\citep{zhang2026memrl}. More recently, RL has been applied at various stages of agent skill development. Specifically, SkillRL~\citep{xia2026skillrl} and D2Skill~\citep{tu2026dynamic} teach smaller models to use skills curated from powerful LLMs in an iterative manner. ARISE~\citep{Li2026ARISEAR} trains a shared policy operating both as skill retriever and worker, with heuristics for skill management. 
Recent studies have begun to train agents for memory or skill curation~\citep{DBLP:journals/corr/abs-2512-17102, DBLP:journals/corr/abs-2602-10652}, but their supervision is mostly restricted to local adaptation within short task streams. This favors immediately useful operations such as skill insertion, while offering limited signal for complex management operations, such as revising outdated skills and deleting harmful ones. \ours{} instead formulates skill curation as a long-horizon, executor-grounded learning problem. We group related tasks into training instances and combine downstream task outcomes with intermediate rewards, turning delayed and indirect feedback into learning signals for skill curation.
\section{Methodology}

In this section, we first formalize the problem setting and introduce the multi-agent modular design of \ours{}. We then detail the RL training recipe designed specifically for training the skill curator.

\subsection{Streaming Skill Curation with Multi-Agent Modular Design}

We consider a \textit{streaming} test-time setting~\citep{wu2024streambench}, where an LLM-based agent is deployed to solve a sequence of tasks
$\mathcal{D}=\{x_1,x_2,\dots,x_T\}$ that arrive over time. 
At each time stamp $t$, the agent must solve the current task $x_t$ before observing future tasks, producing an execution trajectory
$\xi_t=\{o_1,a_1,\dots,o_n,a_n\}$, where $o$ and $a$ denote observations and actions, respectively.
This setting naturally captures the challenge of self-evolving agents, where the system must distill useful experience from the trajectories of past interactions to improve performance on future tasks, and become more capable over time. Figure~\ref{fig: skillos_overview}\textcolor{redlinkcolor}{(a)} presents an overview of the system.

\noindent\textbf{Skill Repository.}
We maintain an external skill repository $\mathcal{S}_t$ at time stamp $t$, which consists of $N_t$ reusable skills
$\mathcal{S}_t=\{s_t^{1},s_t^{2},\dots,s_t^{N_t}\}$.
Following the widely adopted \texttt{SKILL.md} format~\citep{anthropic_skills_2025}, each skill is represented as a single Markdown file with two components as shown in Figure~\ref{fig: skillos_overview}\textcolor{redlinkcolor}{(b)}: 
(i) \textit{YAML frontmatter}, which specifies the skill name and a natural-language description of when the skill should be used, and 
(ii) \textit{Markdown instructions}, which describe the executable knowledge, workflows, constraints, and reusable heuristics captured by the skill.

\noindent\textbf{Agent Executor.}
Given a task $x_t$, a frozen agent executor $\pi_{\mathcal{L}}$ solves the task conditioning on the current environment observation and relevant skills.
Specifically, we retrieve a subset of skills $\tilde{\mathcal{S}}_{t}\subseteq \mathcal{S}_{t}$ using BM25~\citep{robertson2009probabilistic} for each task $x_t$, and the executor samples actions following
$a \sim \pi_{\mathcal{L}}(\cdot \mid x_t, o_t, \tilde{\mathcal{S}}_{t})$.

\noindent\textbf{Skill Curator.}
After the executor completes task $x_t$, the skill curator $\pi_{\mathcal{S}}$ observes the trajectory $\xi_t$, the self-judged correctness of the answers/interactions $\mathbbm{1}_{\xi_t}$, and a retrieved subset of related skills $\tilde{\mathcal{S}}_{t}$.
It then generates a sequence of structured curation operations
$c_t = (u_t^1,\dots,u_t^{M_t})
    \sim \pi_{\mathcal{S}}(\cdot \mid \xi_t, \mathbbm{1}_{\xi_t}, \tilde{\mathcal{S}}_{t})$,
where each operation $u_t^m$ is one of $\{\insertop, \updateop, \deleteop\}$.
Each operation is implemented as a function call (detailed signature in Figure~\ref{fig: tool_signature}) that manipulates the skill repository $\mathcal{S}_t$.
Applying these operations transforms the repository from $\mathcal{S}_{t}$ to $\mathcal{S}_{t+1}$ as
$\mathcal{S}_{t+1} = \textsc{ApplyOps}(\mathcal{S}_{t}, c_t)$.
The updated repository is then used by the executor on subsequent tasks, forming a closed loop between task execution and experience-driven skill evolution.

\subsection{Learning Skill Curation with RL}

\begin{figure}[t]
\begin{center}
\includegraphics[width=0.98\textwidth]{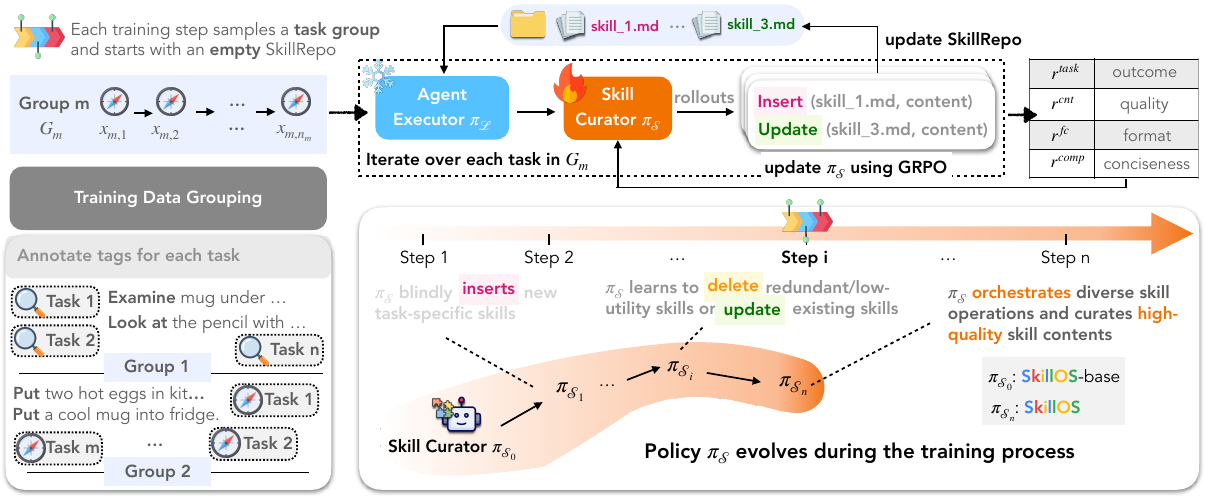}
\end{center}
\vspace{-3mm}
\caption{\ours{} training pipeline. Each training step samples a group of related tasks and initializes an empty \textsc{SkillRepo}. $\pi_{\mathcal{S}}$ is optimized with composite rewards, enabling self-evolution.
}
\label{fig: skillagent}
\vspace{-5mm}
\end{figure}

We optimize the skill curator $\pi_{\mathcal{S}}$ with RL and keep the agent executor $\pi_{\mathcal{L}}$ frozen. 
The main challenge is indirect and delayed feedback for curation decisions, which is only revealed through $\pi_{\mathcal{L}}$'s performance on future relevant tasks.
We address this by constructing grouped training instances (\S~\ref{sec:training_instance_construction}) and designing a composite reward (\S~\ref{sec:reward_design}) that combines future task outcomes with intermediate signals on operation validity, skill quality, and the conciseness of skills. An overview of the training process is shown in Figure~\ref{fig: skillagent}.

\subsubsection{Training Instance Construction}\label{sec:training_instance_construction}

To provide downstream learning signals for skill curation, we construct each training instance as a group of related tasks that are solved sequentially. Within each group, \textsc{SkillRepo} is updated by the curator $\pi_{\mathcal{s}}$ after each task, allowing skills derived from earlier experiences to be evaluated by whether they help solve related future tasks.
This also differs from prior work that focuses on short-horizon transfer~\citep{DBLP:journals/corr/abs-2512-17102, DBLP:journals/corr/abs-2602-10652}, where our grouped formulation exposes the curator to longer skill-evolution trajectories and provides denser feedback for learning complex curation operations.  

Concretely, for each task $x_i$ in $\mathcal{D}=\{x_i\}_{i=1}^{N}$, we first annotate each instance with a set of skill-relevant attributes. Formally, for each $x_i$, we use Gemini-2.5-Pro~\citep{DBLP:journals/corr/abs-2507-06261} to produce a set of tags:
\vspace{-1mm}
\[
Z_i = \{z_i^{1}, z_i^{2}, \dots, z_i^{|Z_i|}\},
\]
where each attribute $z_i$ captures a salient aspect of the task $x_i$, such as topic and common pitfalls. For example, in mathematical reasoning, attributes may include labels such as ``algebra'' or ``Fourier transformation''. These attributes serve as proxies for task-relatedness and potential skill dependency.

Based on the annotated attributes, we then partition $\mathcal{D}$ into a collection of $M$ task groups using the similarity of attributes of these data samples:
\vspace{-1mm}
\[
\mathcal{D}=\{G_1, G_2, \dots, G_{M}\}, \qquad G_m = \{x_{m,1}, x_{m,2}, \dots, x_{m,|G_m|}\},
\]
where all instances within the same group $G_m$ exhibit non-trivial dependency in terms of required skills. Detailed description of data processing and grouping algorithms can be found in Appendix~\ref{app:group_training}.

\subsubsection{Training Loop and Policy Optimization}\label{sec:reward_design} 

We employ Grouped Reward Policy Optimization (GRPO~\cite{DBLP:journals/corr/abs-2402-03300}) for its training stability and sample efficiency. The training loop shown in Algorithm~\ref{alg:mem_grpo} optimizes the skill curator policy $\pi_{\mathcal{S}}$ to maximize a composite reward function over the distribution of generated traces. 
For a task group $G=(x_1,\dots,x_{|G|})$, the curator produces a sequence of curation decisions
$c=(c_1,\dots,c_{|G|})$ as the executor proceeds through the group. Each training step, the reward combines four signals:
\setlength{\abovedisplayskip}{0pt}
\setlength{\belowdisplayskip}{0pt}
\setlength{\abovedisplayshortskip}{0pt}
\setlength{\belowdisplayshortskip}{0pt}
\begin{equation}
    r \;=\; \underbrace{r^{\text{task}}}_{\text{task outcome}}
    + \;\lambda_{\mathrm{f}} \underbrace{r^{\text{fc}}}_{\text{function call}}
    + \;\lambda_{\mathrm{u}} \underbrace{r^{\text{cnt}}}_{\text{content quality}}
    + \;\lambda_{\mathrm{c}} \underbrace{r^{\text{comp}}}_{\text{compression}}
    \label{eq:reward}
\end{equation}

\noindent\textbf{Task outcome reward.}\;
The first task uses an empty \textsc{SkillRepo}, before any curator update occurs.
We thus define the task outcome reward as the average success over the remaining tasks as $r^{\text{task}}
    =
    \frac{1}{|G|-1}
    \sum_{i=2}^{|G|}
    \mathbbm{1}(\xi_i)$,
which provides executor-grounded signal on downstream performance achieved by the evolving \textsc{SkillRepo} from $\pi_{\mathcal{S}}$.

\noindent\textbf{Function call reward.}\;
The function call reward measures whether the curator produces valid skill operations. 
For each curation decision $c_i$, let $\mathrm{Valid}(c_i)$ be the fraction of generated function calls that are valid and successfully executed.
We define the function call reward as $r^{\text{fc}}
    =
    \frac{1}{|G|}
    \sum_{i=1}^{|G|}
    \mathrm{Valid}(c_i)$.

\begin{algorithm}[t]
\caption{Training Skill Curator with Task Groups using GRPO}
\label{alg:mem_grpo}
\begin{algorithmic}[1]
\For{each training step}
    \State $G=(x_1,\dots,x_{|G|})$,\ $\mathcal{S}\leftarrow \emptyset$
        \Comment{\textcolor{googleblue}{\textit{Sample a task group and initialize SkillRepo}}}
    \For{task index $i = 1,\dots,|G|$}
        \State $\tilde{\mathcal{S}} \leftarrow \textsc{BM25}\!\left(x_i,\; \mathcal{S}\right)$
            \Comment{\textcolor{googleblue}{\textit{Retrieve relevant skills}}}
        \State $\xi_{i} \leftarrow \textsc{RunTask}\!\left(\tilde{\mathcal{S}},\;\pi_{\mathcal{L}},\;x_i\right)$
            \Comment{\textcolor{googleblue}{\textit{Run inference on frozen executor}}}
        \State $c_{i} \sim \pi_{\mathcal{S}}\!\left(\cdot \;\middle|\; \xi_i, \tilde{\mathcal{S}}\right)$
            \Comment{\textcolor{googleblue}{\textit{Sample a rollout from skill curator}}}
        \State $\mathcal{S} \leftarrow \textsc{ApplyOps}\!\left(\mathcal{S},\; c_{i}\right)$
            \Comment{\textcolor{googleblue}{\textit{Apply \texttt{insert}/\texttt{update}/\texttt{delete}}}}
    \EndFor
    \State $r \leftarrow \textsc{CalculateReward}(\xi, c)$
    \State $\textsc{Update}\ \pi_{\mathcal{S}}$
        \Comment{\textcolor{googleblue}{\textit{Update skill curator using GRPO}}}
\EndFor
\end{algorithmic}
\end{algorithm}

\noindent\textbf{Compression reward.}\;
To discourage verbatim trajectory copying, we reward concise repository updates. 
Let $\mathcal{S}_i$ denote the skill repository after applying $c_i$, and let $\chi_i$ denote the curator input context at position $i$.
We define $r^{\text{comp}}
    =
    \frac{1}{|G|}
    \sum_{i=1}^{|G|}
    \left(
    1 - \frac{|\mathcal{S}_i|}{|\chi_i|}
    \right)$,
where $|\mathcal{S}_i|$ and $|\chi_i|$ denote token lengths.
This encourages the curator to distill reusable skills rather than store raw trajectories.

\noindent\textbf{Content quality reward.}\;
The content quality reward evaluates whether the curated skills are semantically meaningful and likely to be useful for future tasks. 
Let $\mathrm{Judge}(c_i)$ denote the scalar score assigned by an external judge (Qwen3-32B) $c_i$, we compute the reward as $
    r^{\text{cnt}}
    =
    \frac{1}{|G|}
    \sum_{i=1}^{|G|}
    \mathrm{Judge}(c_i)$.

For each task group $G$, we sample $N$ independent rollouts of the \emph{entire curation sequence} from $\pi_{\mathcal{S}}$. Within each rollout, the executor produces trajectory $\xi_i$ using the skill repository $\mathcal{S}^{i}$ resulting from previous curations $c_{<i}$ till task position $i$ with the same training task group, so different rollouts evolve different repository histories. The GRPO advantage is computed as:
$     A^{n} = r^{n} - \frac{1}{N}\sum_{n'=1}^{N} r^{n'},$
where $r^{n}$ is the composite reward (Eq.~\ref{eq:reward}) for the $n$-th rollout.
We optimize $\pi_{\mathcal{S}}$ with a clipped surrogate objective over all curation steps $i = 1, \ldots, |G|$:
\begin{equation}
    \mathcal{L} = \mathbb{E}_{n} \!\left[\min\!\left(\rho^{n}\,A^{n},\; \mathrm{clip}\!\left(\rho^{n},\, 1{-}\epsilon,\, 1{+}\epsilon\right)\!A^{n}\right)\right]
    \label{eq:grpo}
\end{equation}
where $\rho^{n} = \pi_{\mathcal{S}}(c^{n} \mid \chi) \,/\, \pi_{\theta_{old}}(c^{n} \mid \chi)$ is the importance ratio. The advantage $A^{n}$ is assigned uniformly to all tokens in $c^{n}$, and we discard the KL term in GRPO to encourage policy exploration.

\section{Experiments}

We conduct experiments on both multi-turn agentic tasks and single-turn reasoning tasks, in line with prior work~\citep{xia2026skillrl, wei2025evo, DBLP:journals/corr/abs-2602-10652}. We additionally show that the trained skill curator transfers across agent executors and task domains, highlighting its flexibility and generalizability.

\newcommand{\frozenmark}{%
  \raisebox{-0.12em}{\includegraphics[height=0.9em]{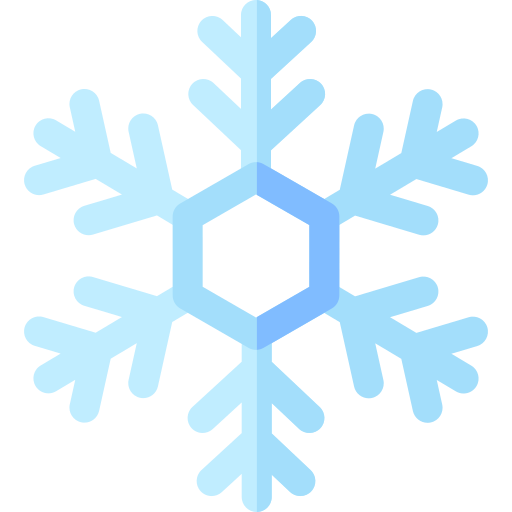}}%
}
\newcommand{\firemark}{%
  \raisebox{-0.12em}{\includegraphics[height=0.9em]{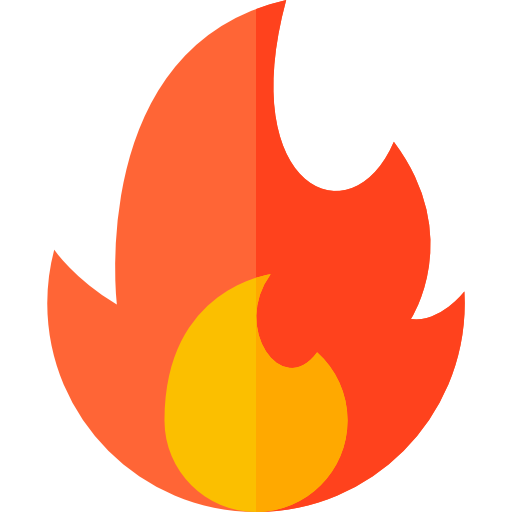}}%
}

\begin{table}[t]
\centering\setlength{\tabcolsep}{3.5pt}
\small
\setlength{\belowcaptionskip}{6.0pt}
  \caption{Experiment results on ALFWorld benchmark. Success rate (SR $\uparrow$) and the number of steps (Steps $\downarrow$) are reported on 6 subsets with 3 different frozen executors.}
  \label{table: alfworld}
  \begin{tabular}{lccccccccc}
    \toprule
    \multirow{2}{*}{{\textbf{Methods}}}& {\textbf{Curator}}&\textbf{Pick}&\textbf{Look}&{\textbf{Clean}}&{\textbf{Heat}}&{\textbf{Cool}}&{\textbf{Pick2}}&{\textbf{{Avg. SR}}}&\multirow{2}{*}{{\textbf{{Steps}}}}\\
    &$\pi_{\mathcal{S}}$&{(35)}&{(13)} &{(27)} & {(16)}&{(25)}&{(24)}&(140)\\
    \midrule
    \multicolumn{10}{c}{\cellcolor{gray!15}\textit{Executor $\pi_{\mathcal{L}}$: Qwen3-8B}} \\
    No Memory&None&\valstd{78.1}{1.6}	&\valstd{46.2}{7.7}	&\valstd{33.3}{13.4}&	\valstd{37.5}{10.8}	&\valstd{29.3}{6.1}&	\valstd{47.2}{6.4}	&\valstd{47.9}{1.2}&	21.1 \\
    ReasoningBank&\frozenmark~Qwen3-8B&\valstd{83.8}{0.0}&	\valstd{48.7}{7.2}&	\valstd{49.4}{16.2}	&\valstd{39.6}{4.4}	&\valstd{41.3}{8.5}	&\valstd{54.2}{8.8}	&\valstd{55.7}{3.1}&	20.1\\
    MemP&\frozenmark~Qwen3-8B&\valstd{80.0}{5.7}&\valstd{43.6}{4.4}&\valstd{24.7}{4.3}&\valstd{33.3}{3.6}&\valstd{38.7}{6.1}&\valstd{48.6}{6.4}&\valstd{49.7}{0.7}&21.0 \\
    \hdashline
    \ours{}-base &\frozenmark~Qwen3-8B&\valstd{79.0}{8.7}&	\valstd{41.0}{4.4}	&\valstd{45.7}{4.3}&	\valstd{37.5}{9.5}	&\valstd{38.7}{4.0}	&\valstd{55.6}{2.1}&	\valstd{53.1}{2.5}	&20.4 \\
    \ours{}-gemini &\frozenmark~Gemini-2.5-Pro&\valstd{77.1}{6.0}&	\valstd{53.8}{6.1}&	\valstd{37.0}{6.4}&	\valstd{37.5}{9.5}	&\valstd{36.0}{3.2}	&\valstd{50.0}{6.7}&	\valstd{50.7}{3.6}	&20.8\\
    \ours{} &\firemark~Qwen3-8B&\valstd{\textbf{85.7}}{3.3}	&\valstd{\textbf{56.4}}{7.7}	&\valstd{\textbf{54.3}}{8.6}	&\valstd{\textbf{43.8}}{9.5}	&\valstd{\textbf{46.7}}{2.3}&	\valstd{\textbf{62.5}}{6.4}	&\valstd{\textbf{61.2}}{4.6}&	\textbf{18.9}\\
    \midrule
    \multicolumn{10}{c}{\cellcolor{gray!15}\textit{Executor $\pi_{\mathcal{L}}$: Qwen3-32B}} \\
    No Memory&None&\valstd{80.0}{2.9}	&\valstd{69.2}{0.0}	&\valstd{45.6}{7.7}	&\valstd{37.5}{16.5}&	\valstd{42.7}{6.1}	&\valstd{43.1}{2.4}	&\valstd{54.5}{2.5}	&20.3\\
    ReasoningBank&\frozenmark~Qwen3-8B&\valstd{86.7}{3.0}	&\valstd{71.8}{5.4}&	\valstd{50.6}{6.3}&	\valstd{45.8}{13.3}&	\valstd{52.0}{8.9}	&\valstd{51.4}{5.1}	&\valstd{61.4}{2.5}&	18.7\\
    MemP&\frozenmark~Qwen3-8B&\valstd{80.0}{2.9}&\valstd{76.9}{0.0}&\valstd{44.4}{7.4}&\valstd{37.5}{10.8}&\valstd{42.7}{2.3}&\valstd{47.2}{6.4}&\valstd{55.7}{3.7}&20.0 \\
    \hdashline
    \ours{}-base&\frozenmark~Qwen3-8B&\valstd{82.9}{2.9}&	\valstd{69.2}{11.8}	&\valstd{48.1}{2.1}&	\valstd{50.0}{9.7}	&\valstd{48.0}{14.4}	&\valstd{52.8}{11.0}&	\valstd{59.8}{3.0}	&19.2\\
    \ours{}-gemini &\frozenmark~Gemini-2.5-Pro&\valstd{\textbf{97.1}}{3.0}&	\valstd{76.9}{5.4}	&\valstd{55.6}{6.0}	&\valstd{43.8}{11.3}&	\valstd{40.0}{5.7}&	\valstd{54.2}{4.9}&	\valstd{63.6}{4.2}&18.1\\
    \ours{} &\firemark~Qwen3-8B&\valstd{91.4}{3.3}&	\valstd{\textbf{76.9}}{4.4}&	\valstd{\textbf{59.3}}{8.6}&	\valstd{\textbf{56.3}}{12.5}&	\valstd{\textbf{57.3}}{10.1}&	\valstd{\textbf{62.5}}{4.2}	&\valstd{\textbf{68.6}}{5.7}	&\textbf{17.3} \\
    \midrule
    \multicolumn{10}{c}{\cellcolor{gray!15}\textit{Executor $\pi_{\mathcal{L}}$: Gemini-2.5-pro}} \\
    No Memory&None&\valstd{90.5}{3.2}&	\valstd{66.7}{5.1}	&\valstd{48.1}{10.2}&	\valstd{39.6}{17.1}	&\valstd{68}{7.4}	&\valstd{68.1}{3.8}	&\valstd{66.4}{2.0}&	17.7\\
    ReasoningBank&\frozenmark~Qwen3-8B&\valstd{91.4}{3.4}&	\valstd{61.5}{4.1}	&\valstd{63.0}{9.3}	&\valstd{39.6}{10.3}	&\valstd{70.7}{3.2}&	\valstd{76.4}{8.5}&	\valstd{71.4}{2.9}	&16.0\\
    MemP &\frozenmark~Qwen3-8B& \valstd{95.2}{2.1}&\valstd{\textbf{74.4}}{6.8}&\valstd{61.7}{7.6}&\valstd{56.3}{12.4}&\valstd{76.0}{6.2}&\valstd{68.1}{8.5}&\valstd{74.3}{3.4}&15.2\\
    \hdashline
    \ours{}-base&\frozenmark~Qwen3-8B&\valstd{91.4}{1.6}	&\valstd{69.2}{7.7}&	\valstd{56.8}{5.7}&	\valstd{54.2}{13.7}&	\valstd{72.0}{4.0}&	\valstd{66.7}{11.0}&	\valstd{70.7}{3.0}&	16.3 \\
    \ours{}-gemini&\frozenmark~Gemini-2.5-Pro&\valstd{94.3}{5.7}&	\valstd{69.2}{0.0}&	\valstd{\textbf{77.8}}{5.7}	&\valstd{\textbf{75.0}}{16.5}&	\valstd{80.0}{12.2}&	\valstd{66.7}{2.4}	&\valstd{79.3}{2.6}&	14.9\\
    \ours{}&\firemark~Qwen3-8B&\valstd{\textbf{95.2}}{2.9}	&\valstd{71.8}{7.7}&\valstd{74.1}{13.0}	&\valstd{72.9}{10.1}&	\valstd{\textbf{77.3}}{6.1}	&\valstd{\textbf{77.8}}{10.0}&	\valstd{\textbf{80.2}}{3.1}&	\textbf{14.8}\\
  \bottomrule
\end{tabular}
\vspace{-5mm}
\end{table}

\subsection{Setup}
We briefly discuss the experiment setup throughout this paper. Full description of datasets, implementations, baselines, and evaluations can be found in Appendix~\ref{app: implementation_details}.

\noindent\textbf{Dataset.} For agentic tasks, we conduct experiments on ALFWorld~\citep{shridhar2021alfworld} and WebShop~\citep{10.5555/3600270.3601778}. ALFWorld is a text-based interactive environment aligned with the ALFRED embodied AI benchmark, where agents must complete household tasks through textual navigation and object manipulation. WebShop simulates an online shopping environment in which agents navigate a realistic web interface to identify and purchase products that satisfy user-specified requirements. For each benchmark, we train \ours{} on its training split where $Z_i$ is the default task type annotations, and evaluate on the corresponding test set. In addition to agentic tasks, we also benchmark for single-turn reasoning tasks, including AIME24, AIME25, and GPQA-Diamond~\citep{rein2024gpqa}. Training data are constructed from DeepMath-103k~\citep{he2026deepmathk}, where we randomly sample a subset of 33,000 data points. 

\noindent\textbf{Evaluation Configurations.} 
We evaluate all methods across two dimensions, \textit{effectiveness} and \textit{efficiency}. For effectiveness, we measure the success rate (SR) and accuracy for agentic tasks and reasoning tasks, respectively. For efficiency, we compute the number of execution steps per agentic task and the number of tokens per reasoning problem, respectively. We compare \ours{} with three categories of baselines: 
(i) a memory-free agent (No Memory); 
(ii) existing memory-based methods, including ReasoningBank~\citep{ouyang2026reasoningbank}, which distills reusable insights from past experiences, and MemP~\citep{DBLP:journals/corr/abs-2508-06433}, which induces procedural memory with advanced memory-management strategies; and 
(iii) internal variants of our framework, including \ours{}-base, which uses the initial skill curator without RL training, and \ours{}-gemini, which uses Gemini-2.5-Pro to directly perform skill curation instead of learning the curator with RL. All prompts used can be found in Appendix~\ref{app: prompt}.

\noindent\textbf{Implementation Details.} 
We use Qwen3-8B~\citep{DBLP:journals/corr/abs-2505-09388} as the base model for $\pi_{\mathcal{S}}$. The frozen executor is also instantiated with Qwen3-8B during training. We train our model using GRPO with a learning rate $1\times10^{-6}$, batch size $32$, and group size $8$. Training is conducted on 16 H100 GPUs using the \texttt{verl} framework~\citep{sheng2024hybridflow}. The full training process takes approximately 3 days for ALFWorld, 2.5 days for reasoning tasks, and 5 days for WebShop.
For testing, we additionally include Qwen3-32B, Gemini-2.5-Pro~\citep{DBLP:journals/corr/abs-2507-06261}, and Gemini-3.1-Flash-Lite (Appendix~\ref{sec:appendix-gemini-flash}) as executors to evaluate the generalization of \ours{} under different executor scales and architectures. Task outcome signal $\mathbbm{1}_{\xi_t}$ is obtained via LLM-as-a-judge with the frozen agent executor (prompt shown in Appendix~\ref{app: prompt}). We use ReAct~\citep{DBLP:conf/iclr/YaoZYDSN023} for agent execution and CoT~\citep{DBLP:conf/nips/Wei0SBIXCLZ22} for reasoning tasks. For the reward function, we set $\lambda_f=1.0$, $\lambda_u=0.1$, and $\lambda_c=0.05$.
We report averaged performance and standard deviation over 3 runs. 

\begin{table*}[t]
\centering
\setlength{\tabcolsep}{5.5pt}
\small
\caption{Experiment results on WebShop and single-turn reasoning tasks for 3 different frozen executors. For WebShop, the averaged score, success rate (SR $\uparrow$), and the number of steps (Steps $\downarrow$) are reported. For reasoning tasks, accuracy (Acc. $\uparrow$) is reported on three datasets.}
\label{tab:merged_results}

\begin{tabular}{lcccccccc}
    \toprule
    \multirow{2}{*}{\textbf{Methods}} 
    & \textbf{Curator}
    & \multicolumn{3}{c}{\textbf{WebShop}} 
    & \multicolumn{4}{c}{\textbf{Reasoning}} \\
    \cmidrule(lr){3-5} \cmidrule(lr){6-9}
    & $\pi_\mathcal{S}$
    & \textbf{Score} & \textbf{SR} & \textbf{Steps}
    & \textbf{AIME24} & \textbf{AIME25} & \textbf{GPQA} & \textbf{Avg. Acc} \\
    \midrule

    \multicolumn{9}{c}{\cellcolor{gray!15}\textit{Executor $\pi_{\mathcal{L}}$: Qwen3-8B}} \\
    No Memory & None
    & \valstd{33.3}{0.7} & \valstd{9.8}{0.5} & 20.3
    & \valstd{76.0}{6.9} & \valstd{71.1}{10.7} & \valstd{61.8}{1.1} & \valstd{69.6}{4.7} \\
    ReasoningBank & \adjustbox{valign=c}{\frozenmark} Qwen3-8B
    & \valstd{35.4}{1.1} & \valstd{11.4}{0.9} & 20.5
    & \valstd{75.4}{5.0} & \valstd{73.2}{10.8} & \valstd{60.3}{3.9} & \valstd{69.6}{2.5} \\
    MemP & \adjustbox{valign=c}{\frozenmark} Qwen3-8B
    & \valstd{35.7}{0.9} & \valstd{12.0}{0.5} & 21.3
    & \valstd{75.6}{5.1} & \valstd{71.1}{5.1} & \valstd{60.6}{4.0} & \valstd{69.1}{4.0} \\
    \hdashline
    \ours{}-base &\adjustbox{valign=c}{\frozenmark} Qwen3-8B
    & \valstd{38.6}{0.9} & \valstd{13.6}{0.8} & 20.1
    & \valstd{75.6}{5.1} & \valstd{71.9}{6.9} & \valstd{59.3}{2.5} & \valstd{68.9}{2.6} \\
    \ours{}-gemini & \adjustbox{valign=c}{\frozenmark} Gemini-2.5-pro
    & \valstd{38.1}{1.0} & \valstd{13.2}{0.9} & 19.6
    & \valstd{73.3}{1.3} & \valstd{71.3}{1.9} & \valstd{57.6}{2.8} & \valstd{67.4}{0.8} \\
    \ours{} & \adjustbox{valign=c}{\firemark} Qwen3-8B
    & \valstd{\textbf{40.6}}{0.7} & \valstd{\textbf{16.5}}{0.7} & \textbf{19.4}
    & \valstd{\textbf{80.0}}{3.3} & \valstd{\textbf{76.7}}{5.8} & \valstd{\textbf{64.6}}{1.3} & \valstd{\textbf{73.8}}{1.8} \\

    \midrule
    \multicolumn{9}{c}{\cellcolor{gray!15}\textit{Executor $\pi_{\mathcal{L}}$: Qwen3-32B}} \\
    No Memory & None
    & \valstd{41.5}{0.5} & \valstd{12.2}{0.3} & 17.0
    & \valstd{81.4}{1.3} & \valstd{72.2}{3.8} & \valstd{68.4}{2.0} & \valstd{74.0}{1.9} \\
    ReasoningBank & \adjustbox{valign=c}{\frozenmark} Qwen3-32B
    & \valstd{40.4}{0.8} & \valstd{11.2}{1.1} & 17.9
    & \valstd{81.1}{9.6} & \valstd{75.6}{5.9} & \valstd{66.9}{1.2} & \valstd{74.9}{2.2} \\
    MemP & \adjustbox{valign=c}{\frozenmark} Qwen3-32B
    & \valstd{30.7}{0.7} & \valstd{10.1}{0.6} & 17.4
    & \valstd{82.2}{5.1} & \valstd{76.7}{0.0} & \valstd{66.5}{2.3} & \valstd{75.1}{2.1} \\
    \hdashline
    \ours{}-base & \adjustbox{valign=c}{\frozenmark} Qwen3-8B
    & \valstd{43.4}{0.8} & \valstd{12.3}{1.0} & 16.8
    & \valstd{80.0}{3.3} & \valstd{75.6}{10.2} & \valstd{67.7}{1.5} & \valstd{74.7}{3.3} \\
    \ours{}-gemini & \adjustbox{valign=c}{\frozenmark} Gemini-2.5-pro
    & \valstd{45.2}{1.0} & \valstd{13.2}{1.1} & 16.6
    & \valstd{77.8}{6.7} & \valstd{74.4}{1.9} & \valstd{66.2}{0.6} & \valstd{73.2}{2.6} \\
    \ours{} &\adjustbox{valign=c}{\firemark} Qwen3-8B
    & \valstd{\textbf{49.2}}{1.2} & \valstd{\textbf{16.5}}{0.6} & \textbf{15.9}
    & \valstd{\textbf{85.6}}{1.9} & \valstd{\textbf{81.1}}{3.3} & \valstd{\textbf{72.4}}{3.0} & \valstd{\textbf{79.7}}{1.6} \\

    \midrule
    \multicolumn{9}{c}{\cellcolor{gray!15}\textit{Executor $\pi_{\mathcal{L}}$: Gemini-2.5-pro}} \\
    No Memory & None
    & \valstd{48.6}{0.3} & \valstd{38.4}{0.5} & 19.5
    & \valstd{85.6}{1.9} & \valstd{80.0}{6.7} & \valstd{79.9}{1.5} & \valstd{81.8}{2.8} \\
    ReasoningBank & \adjustbox{valign=c}{\frozenmark} Gemini-2.5-pro
    & \valstd{50.8}{1.5} & \valstd{40.2}{1.3} & 19.2
    & \valstd{85.6}{5.1} & \valstd{84.4}{6.7} & \valstd{80.4}{2.1} & \valstd{83.5}{2.1} \\
    MemP & \adjustbox{valign=c}{\frozenmark} Gemini-2.5-pro
    & \valstd{51.3}{1.2} & \valstd{39.8}{1.0} & 19.4
    & \valstd{83.3}{6.9} & \valstd{76.7}{5.8} & \valstd{81.8}{3.4} & \valstd{80.6}{3.2} \\
    \hdashline
    \ours{}-base & \adjustbox{valign=c}{\frozenmark} Qwen3-8B
    & \valstd{52.8}{1.0} & \valstd{39.6}{0.8} & 19.0
    & \valstd{87.8}{3.3} & \valstd{83.3}{1.9} & \valstd{82.8}{2.7} & \valstd{84.6}{1.8} \\
    \ours{}-gemini & \adjustbox{valign=c}{\frozenmark} Gemini-2.5-pro
    & \valstd{54.7}{1.0} & \valstd{41.0}{1.2} & \textbf{17.8}
    & \valstd{90.0}{5.1} & \valstd{85.6}{7.7} & \valstd{80.7}{5.5} & \valstd{85.4}{3.5} \\
    \ours{} & \adjustbox{valign=c}{\firemark} Qwen3-8B
    & \valstd{\textbf{56.0}}{0.7} & \valstd{\textbf{41.3}}{0.8} & 18.3
    & \valstd{\textbf{92.2}}{2.4} & \valstd{\textbf{86.7}}{3.5} & \valstd{\textbf{86.8}}{2.1} & \valstd{\textbf{88.6}}{1.5} \\
    \bottomrule
\end{tabular}

\vspace{-5mm}
\end{table*}
\subsection{Main Results}

Tables~\ref{table: alfworld} and~\ref{tab:merged_results} summarize the results for different benchmarks with Qwen3-8B as the skill curator on various agent executors. Based on the results, we have the following observations.

\noindent\textbf{\ours{} achieves strong performance gains across benchmarks.}
Across all three benchmarks, \ours{} consistently outperforms both memory-free and memory-based baselines, showing that the gains come from \emph{learning to manage and evolve} skills rather than from maintaining a static collection. On ALFWorld, \ours{} improves the average success rate from 55.7 to 61.2 over the strongest baseline ReasoningBank with Qwen3-8B as the executor; similar trends hold on WebShop and reasoning tasks. Strikingly, our RL-trained 8B curator even surpasses \ours{}-gemini, despite the latter using a far stronger frontier model as the curator, demonstrating that targeted training of a small curator can outweigh raw model scale. The benefits brought by RL training are also compounded with executor capacity, yielding $+9.5$ absolute improvement with Gemini-2.5-Pro versus $+7.9$ with Qwen3-8B for ALFworld, compared with \ours{}-base.

\noindent\textbf{\ours{} is more efficient, requiring fewer interactions and lower execution cost.}
The gains of \ours{} are accompanied by better efficiency rather than longer trajectories. On ALFWorld, \ours{} reduces the average interaction steps by $2.2$, $3.0$, and $3.1$ compared with ``no memory'' setting with 3 executors, consistently outperforming all memory-based baselines. This trend extends to WebShop, where \ours{} secures higher success rates with fewer environment interactions. These results indicate that the learned skill manager enables the executor to identify procedural shortcuts and bypass redundant exploration. Rather than relying on additional trial-and-error, \ours{} improves performance by distilling experience into direct, actionable expertise that simplifies task execution.

\noindent\textbf{The gains differ between agentic and reasoning tasks, reflecting different forms of reusable skills.}
A notable trend is that the gains of \ours{} are generally larger on multi-turn agentic benchmarks than on single-turn reasoning tasks. We hypothesize that this difference arises from how reusable skills manifest across task types. Agentic tasks naturally expose procedural regularities, such as action ordering, exploration strategies, recovery behaviors, and environment-specific constraints, which can be repeatedly composed and refined across task streams. Reasoning tasks also benefit from skill curation, but their reusable knowledge often appears at a more abstract level, such as decomposition heuristics, constraint formulation, or verification patterns, rather than as directly reusable action procedures. As a result, \ours{} still improves reasoning performance, while the gains are typically smaller than those observed on agentic benchmarks. We provide a case study demonstrating skills curated for different tasks in Figure~\ref{fig: skill_case}.

\subsection{Generalization of \ours{}}\label{exp:generalization}

\begin{figure}[t]
\begin{center}
\includegraphics[width=0.95\textwidth]{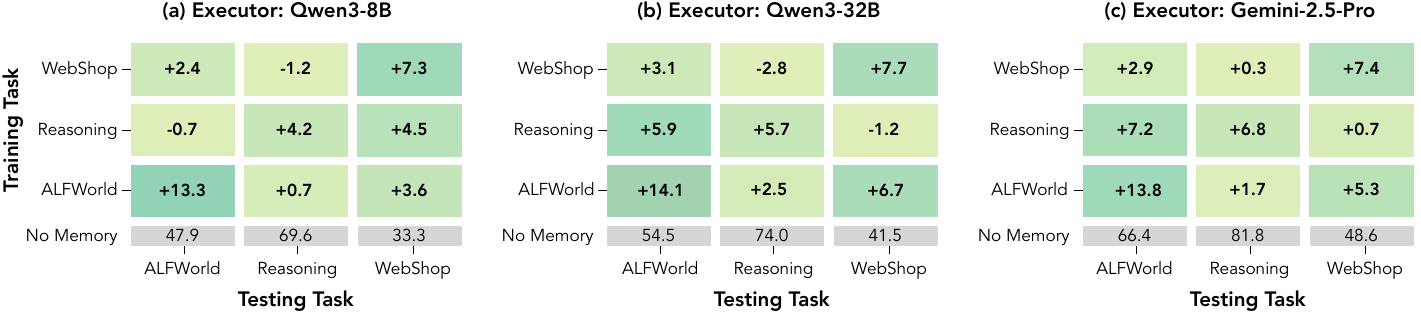}
\end{center}
\vspace{-3mm}
\caption{Cross-task generalization results of \ours{} with (a) Qwen3-8B, (b) Qwen3-32B, and (c) Gemini-2.5-Pro as frozen executors. We plot relative improvement with baselines from \textbf{\textcolor{lightgreen}{least}} to \textbf{\textcolor{deepgreen}{most}}.}
\label{fig: task_generalization}
\vspace{-5mm}
\end{figure}

\noindent\textbf{\ours{} is transferable and remains effective for different agent executors.}
During training, we use Qwen3-8B as the executor. To test whether \ours{} brings improvement for executors that are not seen in training, we pair the trained skill curator with different executors. As shown in Table~\ref{table: alfworld} and \ref{tab:merged_results}, \ours{} consistently improves a wide range of frozen executors across benchmarks, from open-source models (Qwen3-8B, Qwen3-32B) to frontier models (Gemini-2.5-Pro). On ALFWorld, it lifts the average success rate of Qwen3-8B from 47.9 to 61.2 and Gemini-2.5-Pro from 66.4 to 80.2, demonstrating compatibility with executors of varying capacity. Notably, using Gemini-2.5-Pro directly as the curator (\ours{}-gemini) underperforms our trained curator, especially when paired with the smaller Qwen3-8B executor. This highlights a curator-executor mismatch: stronger reasoning ability alone does not guarantee effective skill curation, as frontier-generated skills may be misaligned with the executor's capacity or usage patterns. By contrast, \ours{} learns executor-grounded curation behaviors through RL, producing skills that better match the downstream agent.

\noindent\textbf{\ours{} delivers consistent performance improvement when generalized to different task domains.} Figure~\ref{fig: task_generalization} shows that the skill curator learned by \ours{} transfers well across different tasks. While training and testing on the same task often gives the strongest gain, most off-diagonal entries still bring performance improvement over baselines, indicating that \ours{} captures reusable skills beyond task-specific heuristics. Specifically, skill curator $\pi_{s}$ learned from reasoning tasks transfer particularly well to the two agentic tasks, likely because they contain more abstract and high-level strategies, such as decomposition, verification, and adaptive planning, which are broadly useful across settings. In contrast, skills learned from WebShop or ALFWorld are more tied to environment-specific knowledge, making them less transferable across tasks.
\section{Analysis}

Beyond performance, we analyze \emph{why} \ours{} works, focusing on design choices, evolution of curator's behaviors and contents in \textsc{SkillRepo}, and the role of curated skills in task success. Additional analyses are included in Appendix~\ref{app: additional_analysis}.

\begin{wraptable}{t}{0.36\textwidth}
\centering\setlength{\tabcolsep}{4.5pt}
\small
\setlength{\belowcaptionskip}{5pt}
  \caption{Ablation results of reward design on the ALFWorld dataset.}
  \label{table: ablation}
  \begin{tabular}{lcc}
    \toprule
   \textbf{Methods} &Avg. SR&Steps \\
    \midrule
    \ours-GRPO&61.2&18.9 \\
    \quad w/o $r^{cnt}$&58.6&20.1\\
    \quad w/o $r^{comp}$&60.0&19.3\\
    \hdashline
    \quad w/o grouping &57.3&20.6\\
  \bottomrule
\end{tabular}
\vspace{-5mm}
\end{wraptable}

\vspace{-3mm}
\paragraph{Ablation Studies.}
We ablate two components of \ours{}: (i) auxiliary rewards in Eq.~\ref{eq:reward}, and (ii) grouped task streams in \S~\ref{sec:training_instance_construction}. Experiments are conducted on ALFWorld, with Qwen3-8B used as both the curator and executor. As shown in Table~\ref{table: ablation}, removing either reward component hurts performance. Without the content-quality reward, the success rate drops from 61.2 to 58.6, showing the importance of intermediate supervision for guiding skill updates in a pipelined system. Removing the compression reward causes a smaller but consistent drop, suggesting that concise repositories are easier for the executor to use. The most significant degradation comes from using random task sequences (w/o grouping), which lowers the success rate to 57.3. This highlights the importance of training on grouped task streams, in which curation decisions are learned from their downstream impact on related future tasks.

\begin{wrapfigure}{r}{0.4\textwidth}
\vspace{-3mm}
  \includegraphics[width=0.4\textwidth]{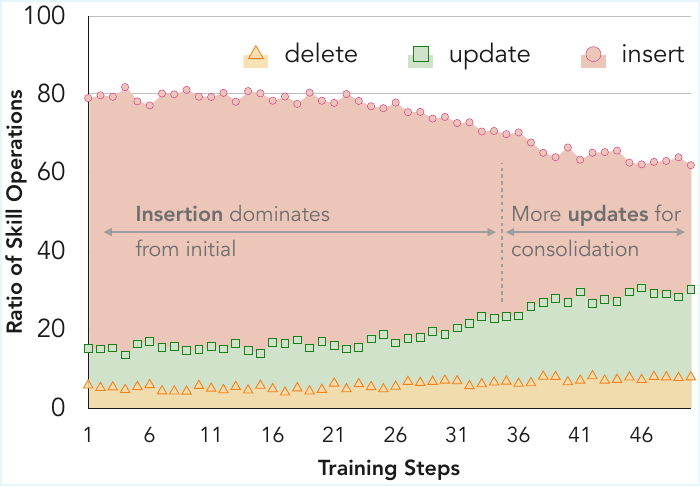}
  \vspace{-5mm}
  \caption{Behaviors of the skill curator w.r.t. skill operations during training.}
  \label{fig: skill-ratio}
  \vspace{-5mm}
\end{wrapfigure}
\noindent\textbf{Behaviors of Skill Curator.}
To better understand how the behavior of the skill curator evolves during training, we analyze the distribution of its three skill operations from rollouts at different training steps: \insertop, \updateop, and \deleteop. Figure~\ref{fig: skill-ratio} plots the proportion of each operation.
At the beginning of training, \texttt{insert} overwhelmingly dominates, indicating that the model is primarily focused on populating the skill repository with new knowledge distilled from experience. As training progresses, however, \texttt{update} becomes increasingly frequent, while \texttt{insert} steadily declines. This suggests that the skill curator gradually moves from plain expansion of skills to refining existing skills. Meanwhile, \texttt{delete} remains a relatively small fraction throughout training with a slightly growing trend, showing the effectiveness of rewarding conciseness of \textsc{SkillRepo}. Instead, the dominant form of adaptation is to revise and consolidate previously acquired skills.

\begin{figure}[H]
\begin{center}
\includegraphics[width=\textwidth]{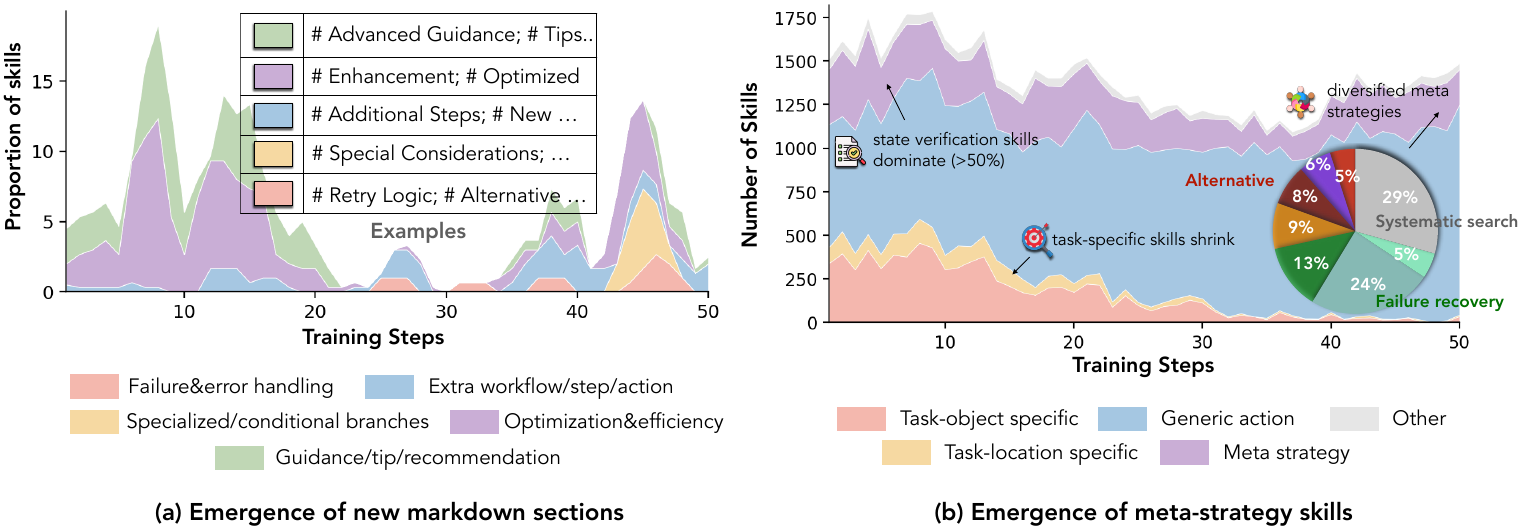}
\end{center}
\vspace{-3mm}
\caption{Evolution dynamics of the curated skills under RL training.}
\label{fig: skill_evol}
\vspace{-5mm}
\end{figure}

\noindent\textbf{Skill Evolution Dynamics.}
Beyond aggregate performance, we examine how the skill repository evolves during RL training. 
We focus on two emergent phenomena: (i) new Markdown sections within individual skills, and (ii) higher-level meta-skills that capture reusable principles across tasks. 
Figure~\ref{fig: skill_evol}(a) shows that early in training, the curator tends to introduce generic sections such as additional guidance, tips, or recommendations, which often make skills more verbose without substantially improving their operational value. 
As training progresses, these additions shift toward more actionable structures, such as failure-handling logic and conditional branches that specify when to deviate from the default workflow. 
This suggests that RL gradually steers the curator from superficial enrichment toward execution-oriented skill refinement.
Figure~\ref{fig: skill_evol}(b) further shows that evolution occurs not only within individual skills, but also in the global organization of the repository. 
Early repositories are dominated by narrow, task-specific skills, whereas later repositories contain a more diverse set of meta-strategy skills covering verification, fallback planning, system search, and strategy adjustment. 
This indicates that the learned curator does not merely accumulate skills, but progressively expands the repository's strategic space, shifting it from isolated task-local procedures toward more compositional cross-task control knowledge.

\begin{wrapfigure}{r}{0.4\textwidth}
\vspace{-3mm}
  \includegraphics[width=0.4\textwidth]{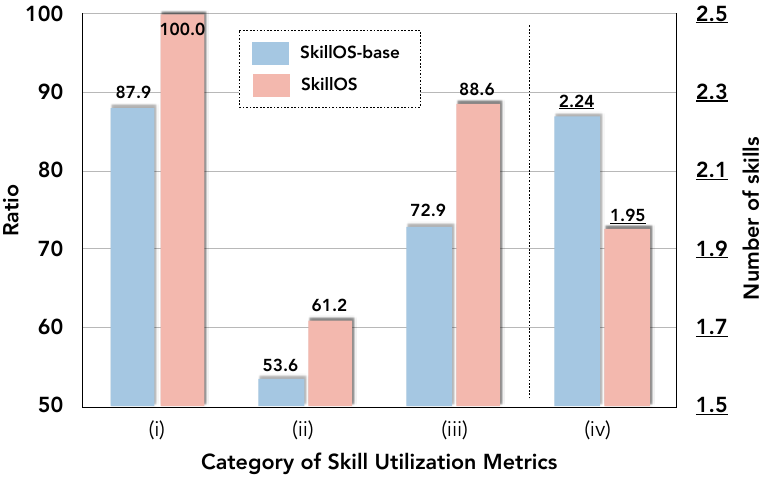}
  \vspace{-5mm}
  \caption{Comparison of skill utilization statistics on ALFWorld.}
  \label{fig: skill-usage}
  \vspace{-5mm}
\end{wrapfigure}

\noindent\textbf{Attribution of Skill Usage.}
To better understand whether the gains of \ours{} come from the evolved skills, we analyze how skills are used during evaluation. 
We consider 4 complementary metrics: (i) \emph{skill usage rate}, the fraction of examples where the agent invokes at least one skill; (ii) \emph{successful skill usage rate}, the success rate among examples that use skills; (iii) \emph{skill coverage}, the fraction of the skill collection that are actually used; and (iv) the \emph{average number of skills used per example}, which measures the degree of skill reliance.
Figure~\ref{fig: skill-usage} reports results on ALFWorld. 
Compared with the baseline, \ours{} invokes skills on \emph{all} evaluation examples and achieves a higher success rate, indicating that the evolved skills contribute directly to task solving. 
Also, a larger fraction of the skill curated by \ours{} is used, showing that RL training improves the overall utility of the curated \textsc{SkillRepo}.
Meanwhile, \ours{} uses fewer skills per example, suggesting that gains come from more precise skill selection rather than more skill context.

\section{Conclusion}

We presented\skillos, an RL training recipe for learning skill curation in self-evolving agents. By decoupling the \emph{skill curator} from the \emph{agent executor}, \ours{} enables modular skill curation without retraining the underlying executor. Through grouped task streams and executor-grounded rewards, \ours{} optimizes curation decisions by their downstream impact on future tasks. Across diverse benchmarks and LLM backbones, \ours{} consistently improves both performance and efficiency. Further analyses show that trained skill curation can outperform frontier models' zero-shot curation ability and generalize across settings, highlighting modular, trained skill curation as a practical path toward agents that self-evolve from experience.

\section{Acknowledgments}
We thank Zilin Xiao, I-Hung Hsu, Zexue He, and members from Google Cloud AI Research for their valuable feedback during the preparation of the paper. Siru was supported by the Molecule Maker Lab Institute: An AI Research Institutes program supported by NSF under Award No. 2019897.

\bibliographystyle{abbrvnat}
\nobibliography*
\bibliography{custom}

\clearpage
\appendix
\newpage
\DoToC

\newpage

\section{Prompts}~\label{app: prompt}
In this section, we provide the full prompt templates used throughout different phases and components of our framework. 

\subsection{Prompt for Skill Curator}

The following prompt templates demonstrate the input to the skill curator during training processes.

\begin{figure}[h]
\begin{center}
\includegraphics[width=\textwidth]{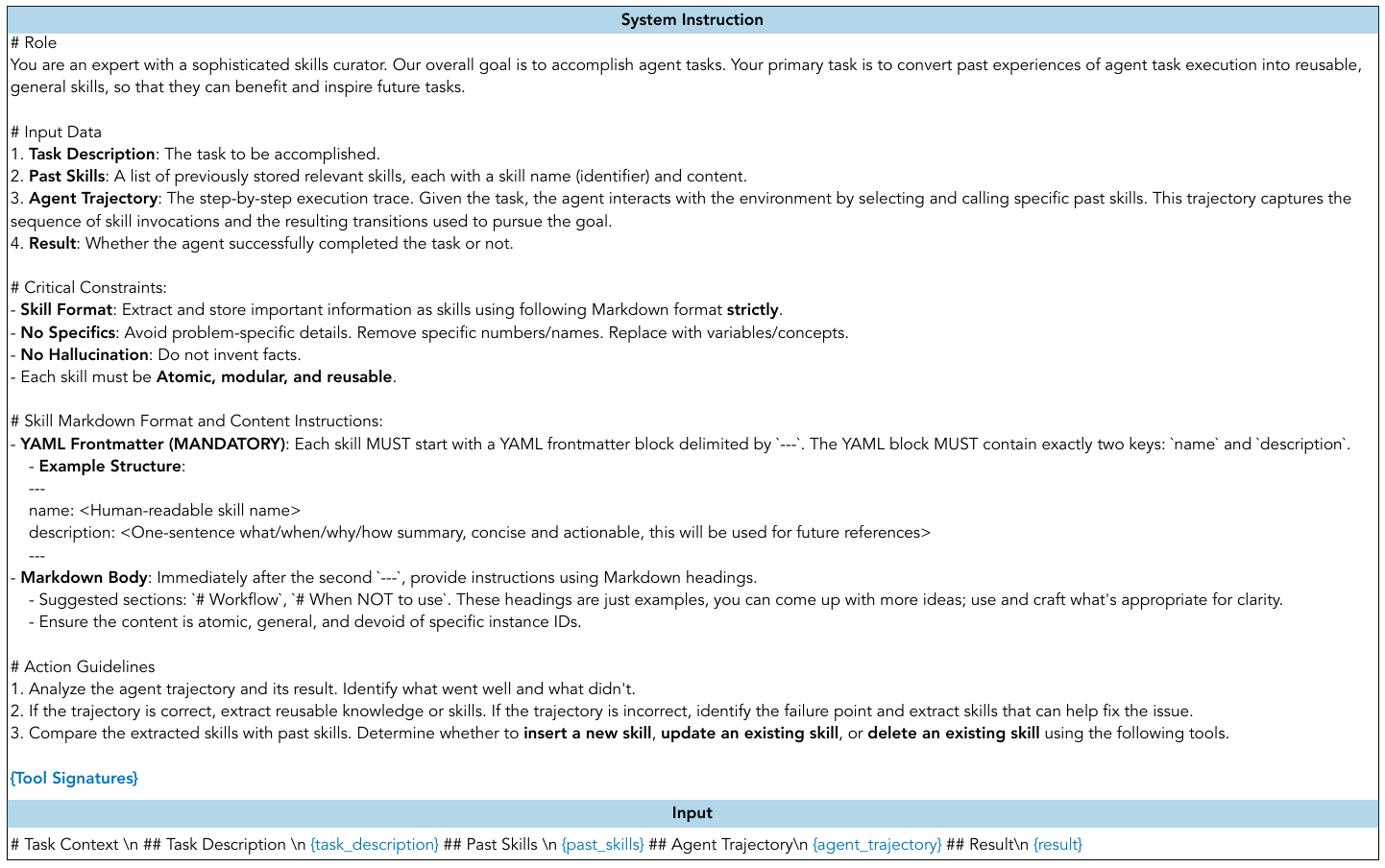}
\end{center}
\vspace{-3mm}
\caption{System prompt used for skill curator during training process.}
\label{fig: skill_curator_prompt}
\vspace{-3mm}
\end{figure}

\begin{figure}[h]
\begin{center}
\includegraphics[width=0.95\textwidth]{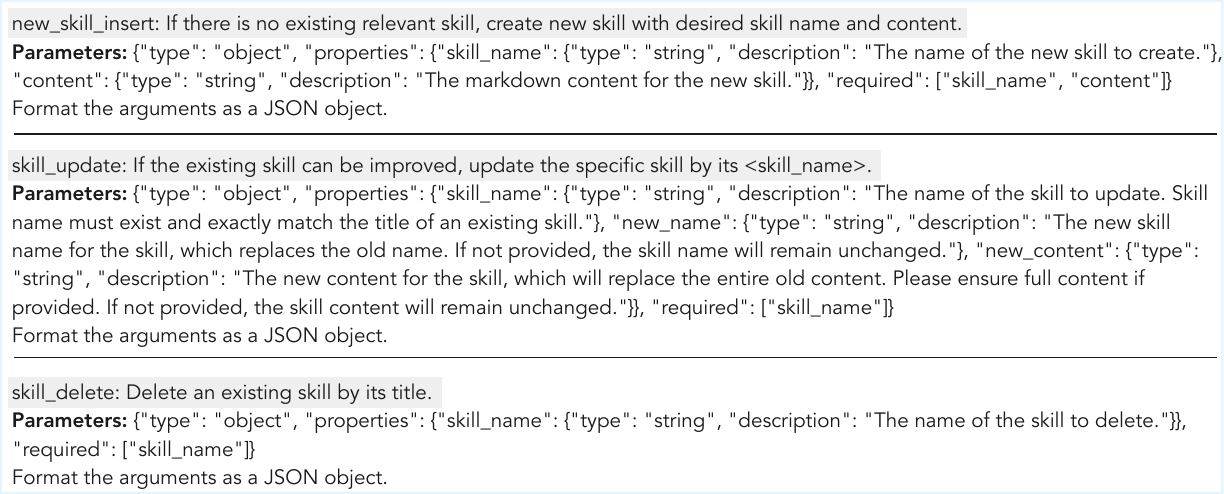}
\end{center}
\vspace{-3mm}
\caption{Tool call definition/signature of skill curator in Figure~\ref{fig: skill_curator_prompt}.}
\label{fig: tool_signature}
\vspace{-3mm}
\end{figure}

\subsection{Prompt for Agent Executor}

The following prompts are used for the frozen agent executor. These templates provide the agent with the current task description, a history of previous interactions, and a set of retrieved skills to guide its decision-making process. All prompts explicitly force chain-of-thought (CoT)~\citep{wei2022chain} reasoning.

For agent tasks including ALFWorld and WebShop, we follow GiGPO~\citep{feng2025group} and leverage its environment and prompt setting for inference. 

\begin{figure}[h]
\begin{center}
\includegraphics[width=\textwidth]{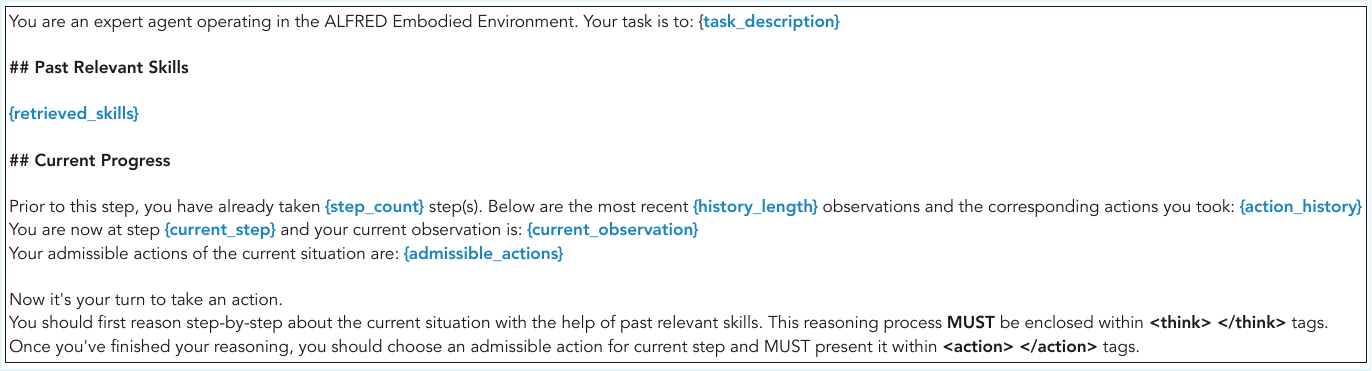}
\end{center}
\caption{Prompt for ALFWorld agent execution with relevant retrieved skills.}
\label{fig: alfworld_system}
\vspace{-5mm}
\end{figure}

\begin{figure}[h]
\begin{center}
\includegraphics[width=\textwidth]{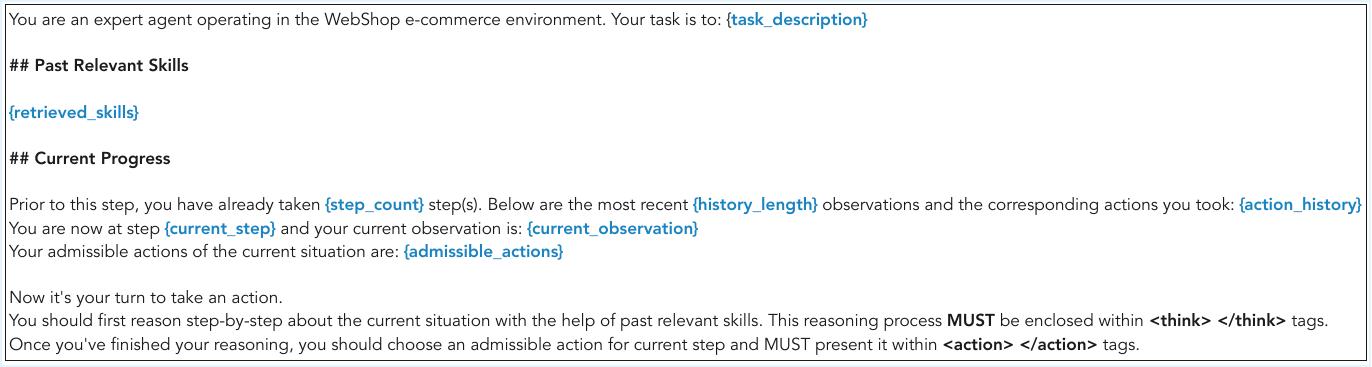}
\end{center}
\caption{Prompt for WebShop agent execution with relevant retrieved skills.}
\label{fig: webshop_system}
\vspace{-5mm}
\end{figure}

\begin{figure}[H]
\begin{center}
\includegraphics[width=0.8\textwidth]{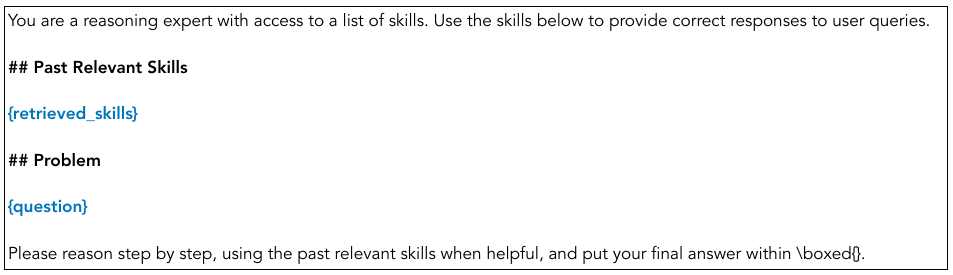}
\end{center}
\caption{Prompt for agent execution in reasoning tasks with relevant retrieved skills.}
\label{fig: reasoning_system}
\vspace{-5mm}
\end{figure}

\subsection{Prompt Used During Training}

During the RL training process, a reward $r^{cnt}$ is assigned based on an external judge of Qwen3-32B to judge whether the curated skills are semantically meaningful and are likely to be useful for future tasks. We show the prompt to the external judge here.

\begin{figure}[H]
\begin{center}
\includegraphics[width=0.9\textwidth]{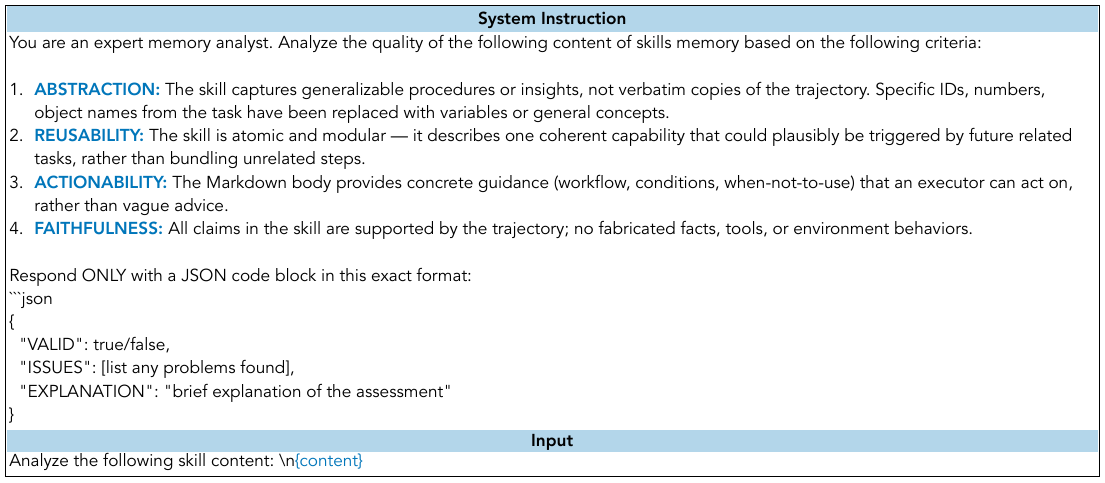}
\end{center}
\caption{Prompt for using an external judge to assign a reward score $r^{cnt}$ for generated skill contents.}
\label{fig: reward_judge_prompt}

\end{figure}

\subsection{Prompt for LLM-as-a-Judge to Obtain Correctness Signals}

\begin{figure}[t]
\begin{center}
\includegraphics[width=0.9\textwidth]{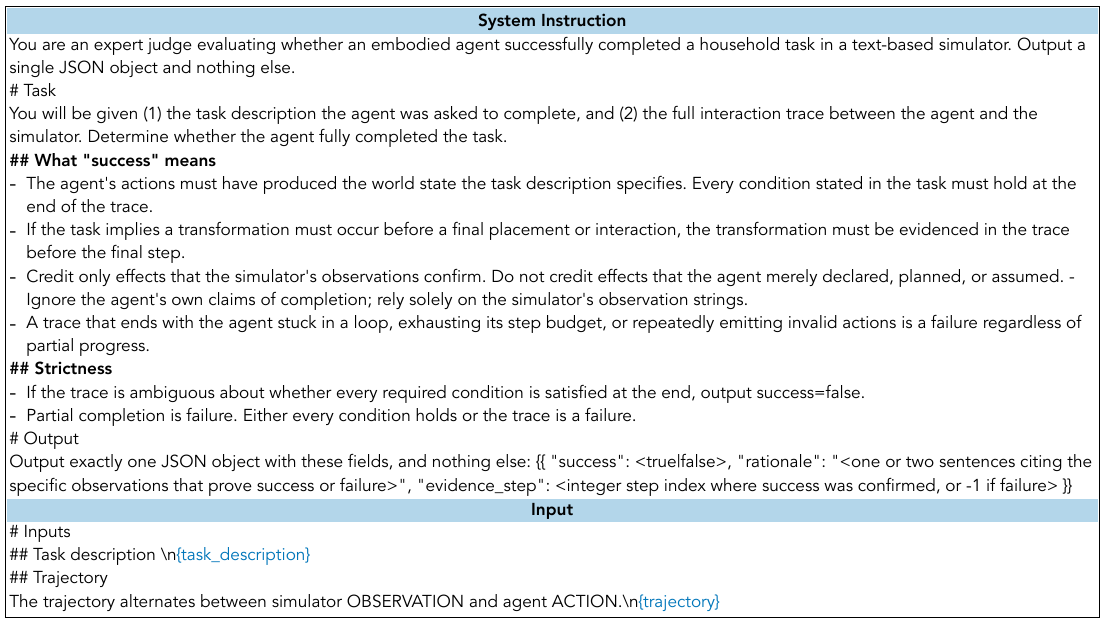}
\end{center}
\caption{Prompt for LLM-as-a-judge to obtain the correctness signal to the current trajectory in the ALFWorld benchmark.}
\label{fig: alfworld_judge}

\end{figure}

\begin{figure}[t]
\begin{center}
\includegraphics[width=0.9\textwidth]{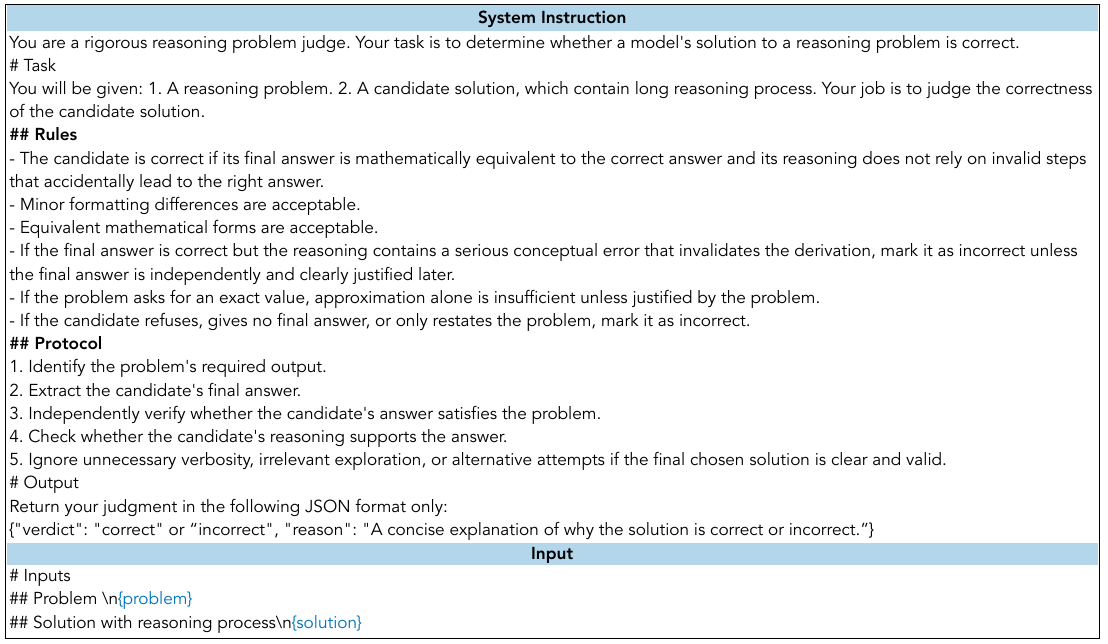}
\end{center}
\caption{Prompt for LLM-as-a-judge to obtain the correctness signal for single-turn reasoning problems.}
\label{fig: reasoning_judge}

\end{figure}

We present the prompts used to obtain the self-judged correctness signal $\mathbbm{1}_{\xi_t}$ for self-evolution via LLM-as-a-judge using the corresponding frozen agent executor as the backbone model in Figures~\ref{fig: alfworld_judge},~\ref{fig: reasoning_judge} for ALFWorld, reasoning, and WebShop tasks, respectively.

\begin{figure}[t]
\begin{center}
\includegraphics[width=0.9\textwidth]{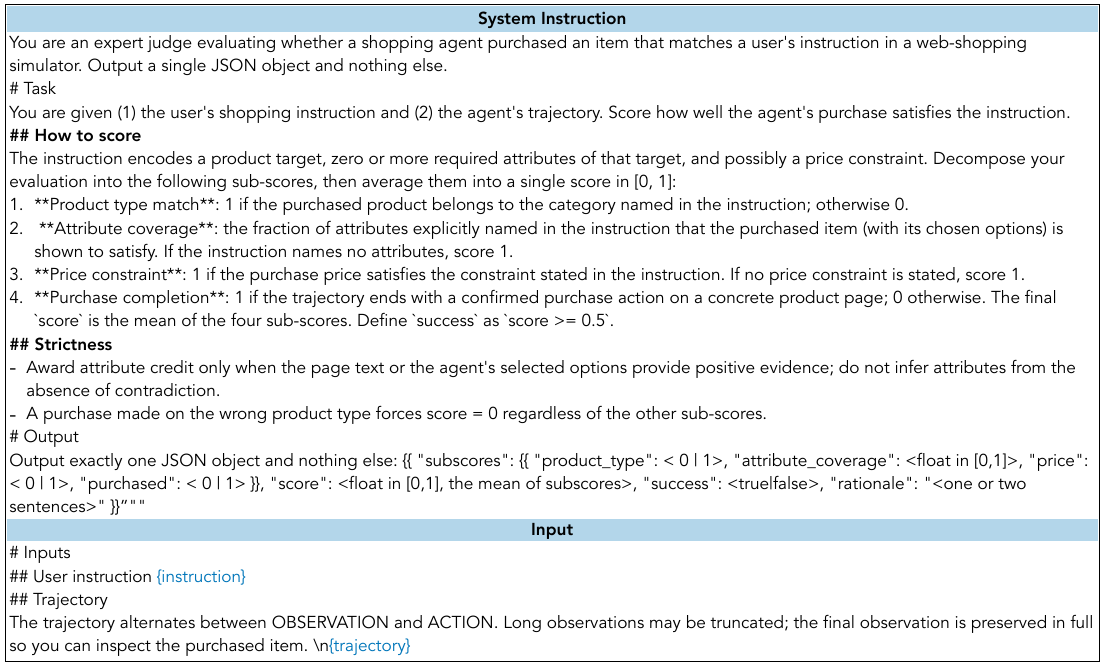}
\end{center}
\caption{Prompt for LLM-as-a-judge to obtain the correctness signal to the current trajectory for the WebShop benchmark.}
\label{fig: webshop_judge}

\end{figure}

\section{Implementation Details}\label{app: implementation_details}

\subsection{Hyperparameters}

We present the choices for all hyperparameters during both the training and inference processes in Table~\ref{tab: hyperparameters} for different tasks.

\begin{table}[h]
    \caption{Hyperparameters for \ours{} for training and inference settings.}
    \begin{center}
    \begin{tabular}{lccc}
        \toprule
        \multirow{2}{*}{Hyperparameter} & \multicolumn{3}{c}{Value} \\
        &ALFWorld&WebShop&Reasoning \\
        \midrule
        \multicolumn{2}{l}{\textit{RL Training}} \\
        Learning rate & \multicolumn{3}{c}{$1 \times 10^{-6}$} \\
        Batch size & \multicolumn{3}{c}{32} \\
        KL loss Coef & \multicolumn{3}{c}{0.001} \\
        Max Prompt Length & \multicolumn{3}{c}{16,384} \\
        Max Response Length & \multicolumn{3}{c}{4,096} \\
        GRPO group size & \multicolumn{3}{c}{8} \\
        Temperature & \multicolumn{3}{c}{1.0} \\
        Steps & 60&50&100 \\
        Data Grouping Size & 10 &10& Random(5,12) \\
        \midrule
        \multicolumn{2}{l}{\textit{Agent Executor Inference}} \\
        Top-K skill retrieval & \multicolumn{3}{c}{5} \\
        Max number of turns&30&30&1 \\
        Action history length & 3&3&-\\
        \bottomrule
    \end{tabular}
    \end{center}
    \label{tab: hyperparameters}
\end{table}

\subsection{Grouping Training Instances}\label{app:group_training}
In this section, we detail the two-stage pipeline used to turn the raw training set $\mathcal{D}=\{x_i\}_{i=1}^N$ into the grouped training set $\mathcal{G}=\{G_j\}_{j=1}^M$ of Section~\ref{sec:training_instance_construction}. Stage~1 annotates each instance with a structured set of latent attributes via an LLM annotator (Sec.~\ref{app:annotation}). Stage~2 assembles groups of related tasks by retrieving, filtering, and ranking candidates under a semantic phrase-level similarity (Sec.~\ref{app:grouping}). For training of single-turn reasoning tasks, we instantiate the pipeline on \textsc{DeepMath-103K}~\citep{he2026deepmathk}, which provides both the raw problems $x_i$ and a scalar difficulty score $d_i\in\mathbb{R}$ that is reused as a curriculum signal by Stage~2. For multi-turn agentic tasks, we leverage the default task type annotation for each benchmark (e.g., 6 task types in ALFWorld) as they naturally expose a discrete partition of tasks into families that share the same underlying skills, and we can use this partition directly in place of the annotated attribute set $Z_i$.
 
\subsubsection{Stage~1: Latent Attribute Annotation}\label{app:annotation}
\begin{figure}[t]
\centering
\begin{center}
    \includegraphics[width=0.9\textwidth]{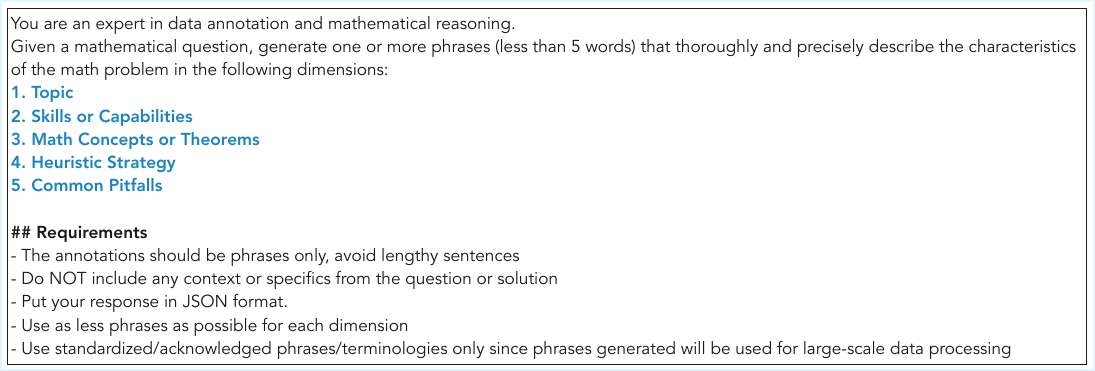}
\end{center}
\caption{System instruction used to elicit $Z_i$ from each task in
$\mathcal{D}$.}\label{fig:annotation_instruction}
\end{figure}

We implement the attribute set $Z_i$ of each instance $x_i$ as a tuple of five phrase-lists,
\[
Z_i \;=\; \bigl(T_i,\; S_i,\; C_i,\; R_i,\; P_i\bigr),
\]
where $T_i$ is the list of high-level \emph{topics}, $S_i$ the required \emph{skills or capabilities}, $C_i$ the underlying \emph{mathematical concepts or theorems}, $R_i$ the applicable \emph{heuristic strategies}, and $P_i$ the \emph{common pitfalls}. Each dimension is populated by a small set of short phrases (at most five words each). The annotator is instructed to: (i)~emit standardized terminology rather than free-form rationales, (ii)~omit any content specific to the question text or its final answer, and (iii)~use as few phrases per dimension as necessary to characterize the task. We enforce the output schema via structured decoding with a fixed JSON response schema, and query Gemini-2.5-Pro with the highest thinking-budget configuration. The exact annotation instruction is reproduced in Figure~\ref{fig:annotation_instruction}.
 
\subsubsection{Stage~2: Group Construction}\label{app:grouping}
Given $\{(x_i,Z_i,d_i)\}_{i=1}^N$, we construct each group $G_j=(x_{j,1},\dots,x_{j,n})$ by sampling a seed task and then iteratively appending related tasks. The core primitive is a pair sampler that, given a source $x_s$, returns an admissible successor $x_t$; longer groups are obtained by iterating this primitive with a growing exclusion set so that instances within a group remain distinct.
 
\paragraph{Phrase similarity.}
Because the annotated phrases come from a large open vocabulary (e.g., \emph{``pigeonhole principle''} vs.\ \emph{``counting argument''}), exact set overlap is unreliable. We therefore score the similarity between any two phrase lists $A$ and $B$ using a \emph{soft-Jaccard} $\mathrm{SJ}_\tau(A,B)$ that combines exact matches with a greedy one-to-one matching between remaining phrases under a sentence-embedding cosine similarity (computed with \texttt{all-MiniLM-L6-v2}~\citep{reimers2019sentence}) above a threshold $\tau$. We write $m_\tau(A,B)$ for the resulting integer \emph{matched-pair count}, which we use alongside $\mathrm{SJ}_\tau$ in the filters below.
 
\paragraph{Dependency gate.}
For a source $x_s$ and candidate $x_t$, we accept the pair only when all of the
following hold:
\begin{enumerate}[leftmargin=1.5em,itemsep=1pt,topsep=1pt]
\item \emph{Shared foundation:} $m_\tau(C_s,C_t)\ge\kappa_C$ and
      $m_\tau(S_s,S_t)\ge\kappa_S$;
\item \emph{Shared reasoning:} $m_\tau(R_s,R_t)+m_\tau(P_s,P_t)\ge 1$;
\item \emph{Not a near-duplicate:} $\mathrm{SJ}_\tau(T_s,T_t)\le\theta_T$ and the weighted
      overall similarity $\Omega(x_s,x_t)\le\sigma_{\max}$;
\item \emph{Not too unrelated:} $\Omega(x_s,x_t)\ge\sigma_{\min}$;
\item \emph{Progression:} $x_t$ introduces at least one new concept or skill,
      i.e.\ $|C_t|>m_\tau(C_s,C_t)$ or $|S_t|>m_\tau(S_s,S_t)$;
\item \emph{Curriculum direction:} $d_t-d_s\ge\delta_{\min}$.
\end{enumerate}
Here $\Omega$ is a convex combination of per-dimension soft-Jaccard scores across
$\{C,S,R,P,T\}$ with weights listed in Table~\ref{tab:group_hparams}. Conditions
(1)--(2) ensure genuine reuse of foundational knowledge and reasoning machinery; (3)--(4)
place the pair in a useful ``related but not redundant'' band; (5) guarantees that
$x_t$ carries something new for the skill curator to compress into the library; and
(6) enforces a forward curriculum.
 
\paragraph{Candidate retrieval and scoring.}
Scoring all $N{-}1$ alternatives per source is prohibitive, so we precompute an inverted index over the dependency fields $\{C,R,P\}$: for each source $x_s$, the candidate pool consists of tasks that share at least one exact dependency phrase with $x_s$, capped at $K_{\text{inv}}$ entries via uniform subsampling. Routing retrieval through dependency fields rather than topics prevents groups from collapsing onto a single narrow subject. Among the candidates that pass the gate, we select the one that maximizes
\[
s(x_s,x_t)\;=\;\sum_{f\in\{C,S,R,P,T\}} w_f\,\mathrm{SJ}_\tau(f_s,f_t)
\;+\;\lambda\cdot b(d_s,d_t),
\]
where $b(\cdot)$ is a bounded difficulty bonus that rewards moderate forward steps. If no inverted-index candidate passes the gate, we fall back to a uniform random pool of size $F$ and re-apply the same gate and scoring; this catches pairs whose phrases agree semantically but not lexically. Extensions sourced from the fallback pool are tagged so downstream training can audit or downweight them. The difficulty gap $d_t-d_s$ is additionally modulated by a randomized curriculum mode $(p_\uparrow,p_=,p_\downarrow)$; for our main experiments, we use an almost exclusively forward curriculum, which produced a more stable training signal than mixed curricula.
 
\paragraph{Hyperparameters.}
Table~\ref{tab:group_hparams} lists all hyperparameters of the Stage~2 pipeline and the values adopted for our main experiments. The weights were tuned on a held-out subset of 200 source tasks by manually inspecting sampled pairs for prerequisite quality; we found the pipeline largely insensitive to small perturbations of the weights but noticeably sensitive to the progression and overall-similarity-band conditions, removing either of which produced markedly more trivial or degenerate pairs.
 
\begin{table}[h]
\centering\small
\caption{Hyperparameters of the Stage~2 grouping pipeline.}\label{tab:group_hparams}
\begin{tabular}{llr}
\toprule
Symbol & Meaning & Value \\
\midrule
---                      & Phrase encoder                                & \texttt{all-MiniLM-L6-v2} \\
$\tau$                   & Cosine threshold for fuzzy phrase matching    & $0.60$ \\
$\kappa_C$               & Minimum matched concept pairs                 & $1$ \\
$\kappa_S$               & Minimum matched skill pairs                   & $1$ \\
$\theta_T$               & Maximum topic soft-Jaccard                    & $0.65$ \\
$\sigma_{\min},\sigma_{\max}$ & Overall-similarity band                  & $0.30,\,0.85$ \\
$\delta_{\min}$          & Difficulty-delta floor                        & $0.0$ \\
$(w_C,w_S,w_R,w_P,w_T)$  & Dimension weights & $(5,\,4,\,3,\,1,\,2)$ \\
$\lambda$                & Difficulty-bonus weight                       & $1.0$ \\
$(p_\uparrow,p_=,p_\downarrow)$ & Mode probabilities                     & $(0.80,\,0.20,\,0.00)$ \\
$[\Delta_{\min},\Delta_{\max}]$ & Gap in \textsc{easy$\rightarrow$hard} mode & $[0.5,\,3.0]$ \\
$\Delta_=$               & Maximum $|d_t-d_s|$ in \textsc{same} mode     & $0.3$ \\
$K_{\text{inv}}$         & Inverted-index subsample cap                  & $2{,}000$ \\
$F$                      & Fallback pool size                            & $200$ \\
\bottomrule
\end{tabular}
\end{table}

\subsection{Experiment Setup}

\subsubsection{Datasets}

In this section, we provide a detailed introduction to all the datasets involved in this paper.

\noindent\textbf{ALFWorld.} ALFWorld~\citep{shridhar2021alfworld} is a text-based interactive benchmark that aligns the TextWorld engine with the embodied ALFRED environment, enabling agents to learn high-level household policies through natural-language interaction. The benchmark covers six task types — Pick \& Place, Examine in Light, Clean \& Place, Heat \& Place, Cool \& Place, and Pick Two \& Place — situated in 120 simulated rooms spanning kitchens, bedrooms, bathrooms, and living rooms. It provides $3,553$ training tasks, together with $140$ valid\_seen tasks for the test set. At each step, the agent receives a textual description of its surroundings together with a goal instruction (e.g., "put a hot apple in the fridge") and must issue high-level commands such as go to, take, open, heat, and put.

\noindent\textbf{WebShop} WebShop~\citep{10.5555/3600270.3601778} is a simulated e-commerce web environment designed to benchmark language agents on realistic, grounded shopping tasks. The environment is populated with 1.18 million real-world products scraped from Amazon and 12,087 crowd-sourced natural-language instructions, partitioned into 10,587 training, 1,000 dev, and 500 test instructions. Given an instruction (e.g., ``I'm looking for a quick-release fitness strap band in teal, priced lower than \$40.00''), the agent interacts with the environment via two action types — search[query] and click[button] — to locate and purchase a product that matches the specified attributes, type, options, and price. At the end of each episode, a programmatic reward in [0, 1] is computed by comparing the purchased item against the ground-truth product specification. Following the standard evaluation protocol used in prior LLM-agent work, we evaluate on the 500 held-out test instructions.

\noindent\textbf{DeepMath-103K} DeepMath-103K~\citep{he2026deepmathk} is a large-scale, decontaminated mathematical reasoning dataset containing approximately 103K problems at high difficulty (primarily AoPS Levels 5–9), spanning algebra, calculus, number theory, geometry, probability, and discrete mathematics. Each problem is paired with a verifiable final answer — enabling rule-based RL rewards — together with a difficulty score, topic label, and three DeepSeek-R1~\citep{guo2025deepseek} chain-of-thought solutions. Specifically, we annotate a subset with around $33,000$ problems, with a final $20,000$ set of grouped training instances.

\noindent\textbf{AIME24 \& AIME25.} A collection of demanding mathematical problems
sourced from the 2024 and 2025 American Invitational Mathematics Examination (AIME), with 30 problems each year. Problems encompass algebra, geometry, number theory, and combinatorics. Created to assess large language models’ sophisticated mathematical reasoning abilities, the dataset presents substantial difficulty, systematic multi-phase solutions, and distinctive answers, establishing it as a robust benchmark for evaluating advanced analytical capabilities.

\noindent\textbf{GPQA.} Short for Graduate Level Google-Proof Q$\&$A Benchmark~\citep{rein2024gpqa}, GPQA comprises a collection of demanding text-based multiple choice problems authored by subject specialists in biology, physics, and chemistry, intentionally crafted to be ``exceptionally challenging''. We use the ``GPQA-Diamond'' subset for testing, which has $198$ problems in total. 

\subsubsection{Baselines}
We compare \ours{} against five representative baselines that span memory-free agents, recent memory-augmented methods, and two internal variants of our own framework. All baselines share the same frozen Agent Executor and are evaluated under identical task suites, retrieval budgets, and decoding settings to isolate the contribution of the memory mechanism.

\noindent\textbf{(i) No Memory.} A memory-free baseline in which the Agent Executor solves each task independently, without access to any external memory or cross-task knowledge transfer. Each episode begins from a blank state, and no information is retained across tasks. This baseline establishes a lower bound and isolates the contribution of any form of accumulated experience.

\noindent\textbf{(ii) ReasoningBank \citep{ouyang2026reasoningbank}.} A recent memory-augmented method that distills reusable reasoning insights from past trajectories and stores them as a searchable bank for future tasks. At inference time, relevant insights are retrieved and injected into the executor's context to guide reasoning. ReasoningBank represents the class of experience-distillation approaches, which emphasize the content of stored knowledge but rely on fixed, heuristic policies for deciding what to write or discard.

\noindent\textbf{(iii) MemP \citep{DBLP:journals/corr/abs-2508-06433}.} A procedural-memory method that induces reusable procedures from agent experience and applies advanced memory-management strategies — including consolidation, forgetting, and re-indexing — to maintain the memory store over time. MemP represents the class of rule-based memory management approaches, which feature more sophisticated maintenance policies than ReasoningBank but still prescribe curation decisions through hand-designed heuristics rather than learning them from downstream task feedback.

\noindent\textbf{(iv) \ours{}-base.} A variant of our framework in which the Skill Curator is instantiated with the same open-source backbone as \ours{} but without any RL fine-tuning, while all other components remain identical to \ours{}. This baseline serves two purposes: (a) it provides a lower-bound reference point that reflects the intrinsic prompting-based curation ability of the open-source backbone prior to optimization, and (b) it isolates the contribution of our GRPO-based training, since \ours{}-base shares exactly the same model architecture, prompting template, and memory interface as \ours{} but forgoes end-to-end optimization against task performance.

\noindent\textbf{(v) \ours{}-gemini.} A variant of our framework in which the Skill Curator is instantiated with Gemini-2.5-Pro instead of a trained open-source model, while all other components remain identical to \ours{}. This baseline serves two purposes: (a) it provides a strong closed-source reference point for the upper bound of prompting-based curation, and (b) it isolates the effect of our GRPO-based training, since \ours{}-gemini shares the same prompting template and memory interface as \ours{} but forgoes RL optimization against task performance.

Together, these baselines cover the main design axes along which memory-augmented agents differ from \ours{}: whether memory exists at all (i), how stored knowledge is represented (ii vs. iii), and whether curation decisions are prescribed by heuristics or learned from task feedback (ii and iii vs. \ours{}), as well as whether the curator itself benefits from RL optimization (iv and v vs. \ours{}).

\subsubsection{Evaluation Metrics}

We evaluate \ours{} and all baselines along two complementary axes --- \textbf{task effectiveness} and \textbf{action efficiency} --- using metrics tailored to each benchmark. Across all benchmarks and methods, every configuration is run with three independent random seeds; we report the mean across seeds, with one standard deviation shown as a subscript (e.g., $85.7_{\pm 1.6}$). Within each backbone block of Tables~\ref{table: alfworld} and~\ref{tab:merged_results}, the best value in each column is highlighted in \textbf{bold}.

\paragraph{Success Rate (SR $\uparrow$).}
Our primary effectiveness metric on both ALFWorld and WebShop. On ALFWorld, SR is the fraction of evaluation episodes in which the agent reaches the goal state within the step budget, yielding a binary $\{0, 1\}$ outcome per episode. We report SR both per task category --- \textit{Pick}, \textit{Look}, \textit{Clean}, \textit{Heat}, \textit{Cool}, and \textit{Pick2} --- and as a macro-average (\textit{Avg.\ SR}) across the six categories, so that categories with fewer tasks are not dominated by larger ones. On WebShop, following~\citep{10.5555/3600270.3601778}, SR is the fraction of episodes whose final reward equals exactly $1$, i.e., the purchased product fully matches all specified attributes, options, type, and price constraints.

\paragraph{WebShop Score ($\uparrow$).}
In addition to SR, WebShop provides a dense per-episode reward in $[0, 100]$ that credits partial matches on attributes, options, type, and price even when the purchase is not a perfect match. We report the average score across evaluation episodes as a finer-grained complement to SR: two methods with similar SR may differ substantially in how close their near-misses are to the target product.

\paragraph{Number of Steps (Steps $\downarrow$).}
Our efficiency metric on ALFWorld and WebShop. \textit{Steps} is the average number of environment actions the agent issues per episode, computed over all evaluation episodes regardless of success. Failed episodes contribute steps up to their termination point (task completion, max-step cutoff, or early stop). This metric captures a dimension that SR and Score alone cannot: two methods may achieve comparable effectiveness while differing substantially in how efficiently they reach the goal, which has direct implications for inference cost and deployment feasibility.

\paragraph{Accuracy (Acc.\ $\uparrow$) on reasoning benchmarks.}
For the single-turn reasoning datasets --- AIME24, AIME25, and GPQA --- we report exact-match accuracy: the fraction of questions whose extracted final answer matches the ground truth. For AIME24 and AIME25, we adopt the evaluation protocol from the HuggingFace \texttt{math\_verify}\footnote{\url{https://github.com/huggingface/Math-Verify}} toolkit, which parses the model's final boxed expression and verifies mathematical equivalence to the reference answer (accounting for equivalent numerical forms, simplifications, and formatting variants). For GPQA, which is a multiple-choice benchmark, we extract the predicted option letter from the model's response and score it as correct if and only if it exactly matches the ground-truth option. We additionally report an average accuracy (\textit{Avg.\ Acc.}) across the three datasets to summarize overall reasoning ability.

\paragraph{Evaluation protocol.}
All methods share the same frozen Agent Executor, retrieval budget (top-$k$ skills retrieved via BM25), maximum step budget, and decoding temperature within each backbone, so that differences in the reported metrics are attributable to the memory mechanism rather than to confounding inference settings. Unless stated otherwise, all numbers in the main paper are computed on the official held-out evaluation splits of each benchmark.

\section{Additional Analyses}\label{app: additional_analysis}

\subsection{Results on Gemini-3.1-Flash-Lite}
\label{sec:appendix-gemini-flash}

In addition to the Qwen3-8B/32B and Gemini-2.5-Pro executors used in the main paper, we further evaluate \ours{} on ALFWorld with the more recent Gemini-3.1-Flash-Lite as the frozen Agent Executor, to verify that our gains generalize to newer model families. Results are reported in Table~\ref{tab:alfworld-flash-lite}.

\ours{} achieves the highest average success rate (73.1\%), outperforming the strongest external baseline ReasoningBank (66.0\%) by \textbf{+7.1 points} and the No-Memory baseline (61.2\%) by \textbf{+11.9 points}, while requiring the fewest interaction steps (15.5 vs.\ 18.5 for No Memory). The two internal variants reproduce the ordering observed in the main experiments: \ours{}-base reaches only 63.6\% — barely above No Memory — confirming that the open-source backbone cannot recover the curation policy through prompting alone, and \ours{}-gemini improves to 71.2\% but is still surpassed by \ours{} despite using a much stronger curator backbone. This reinforces our main finding that \emph{learning} the curator with task-level feedback contributes more than scaling up the curator model. We also note that MemP (58.6\%) underperforms even No Memory under this executor, suggesting that hand-designed curation heuristics are brittle when the executor is less capable, whereas the policy learned by \ours{} remains robust. Per-subset, \ours{} wins on four of six subsets, with particularly large margins on \textit{Look} (84.6\% vs.\ 71.8\%) and \textit{Cool} (68.0\% vs.\ 48.0\%); the remaining two subsets are won by \ours{}-gemini (\textit{Pick}) and ReasoningBank (\textit{Heat}), on which \ours{} nonetheless remains competitive. Overall, these results confirm that the advantage of \ours{} transfers cleanly to a newer executor family.

\begin{table}[t]
\centering\setlength{\tabcolsep}{6.0pt}
\small
\setlength{\belowcaptionskip}{6.0pt}
  \caption{Experiment results on ALFWorld benchmark. Success rate (SR $\uparrow$) and the number of steps (Steps $\downarrow$) are reported on 6 subsets for Gemini-3.1-Flash-Lite as frozen executor.}
  \label{tab:alfworld-flash-lite}
  \begin{tabular}{lcccccccc}
    \toprule
    \multirow{2}{*}{{\textbf{Methods}}} &\textbf{Pick}&\textbf{Look}&{\textbf{Clean}}&{\textbf{Heat}}&{\textbf{Cool}}&{\textbf{Pick2}}&{\textbf{{Avg. SR}}}&\multirow{2}{*}{{\textbf{{Steps}}}}\\
    &{(35)}&{(13)} &{(27)} & {(16)}&{(25)}&{(24)}&(140)\\
    \midrule
    No Memory&\valstd{85.7}{0.0}&	\valstd{59.0}{8.9}	&\valstd{67.9}{9.3}&	\valstd{25.0}{6.2}	&\valstd{38.7}{2.3}	&\valstd{66.7}{0.0}	&\valstd{61.2}{2.3}&	18.5\\
    ReasoningBank&\valstd{87.6}{4.4}&	\valstd{71.8}{4.4}	&\valstd{63.0}{0.0}	&\valstd{\textbf{52.1}}{14.4}	&\valstd{48.0}{10.6}&	\valstd{62.5}{0.0}&	\valstd{66.0}{2.7}	&17.6\\
    MemP & \valstd{84.3}{6.1}&\valstd{57.7}{5.4}&\valstd{63.0}{0.0}&\valstd{28.1}{4.4}&\valstd{34.0}{2.8}&\valstd{62.5}{0.0}&\valstd{58.6}{1.0}&19.3\\
    \hdashline
    \ours{}-base&\valstd{86.7}{1.6}	&\valstd{61.5}{0.0}&	\valstd{66.7}{0.0}&	\valstd{41.7}{6.2}&	\valstd{38.7}{16.0}&	\valstd{68.1}{2.4}&	\valstd{63.6}{3.9}&	17.7 \\
    \ours{}-gemini &\valstd{\textbf{96.2}}{1.6}&	\valstd{61.5}{13.3}&	\valstd{74.1}{3.7}	&\valstd{{31.2}}{12.5}&	\valstd{66.7}{4.6}&	\valstd{68.1}{2.4}	&\valstd{71.2}{2.9}&	16.1\\
    \ours{} &\valstd{88.6}{0.0}	&\valstd{\textbf{84.6}}{13.3}&\valstd{\textbf{77.8}}{0.0}	&\valstd{37.5}{17.2}&	\valstd{\textbf{68.0}}{8.0}	&\valstd{\textbf{68.1}}{2.4}&	\valstd{\textbf{73.1}}{2.7}&	\textbf{15.5}\\
  \bottomrule
\end{tabular}
\end{table}

\subsection{Case Studies}

\begin{figure}[t]
\begin{center}
\includegraphics[width=\textwidth]{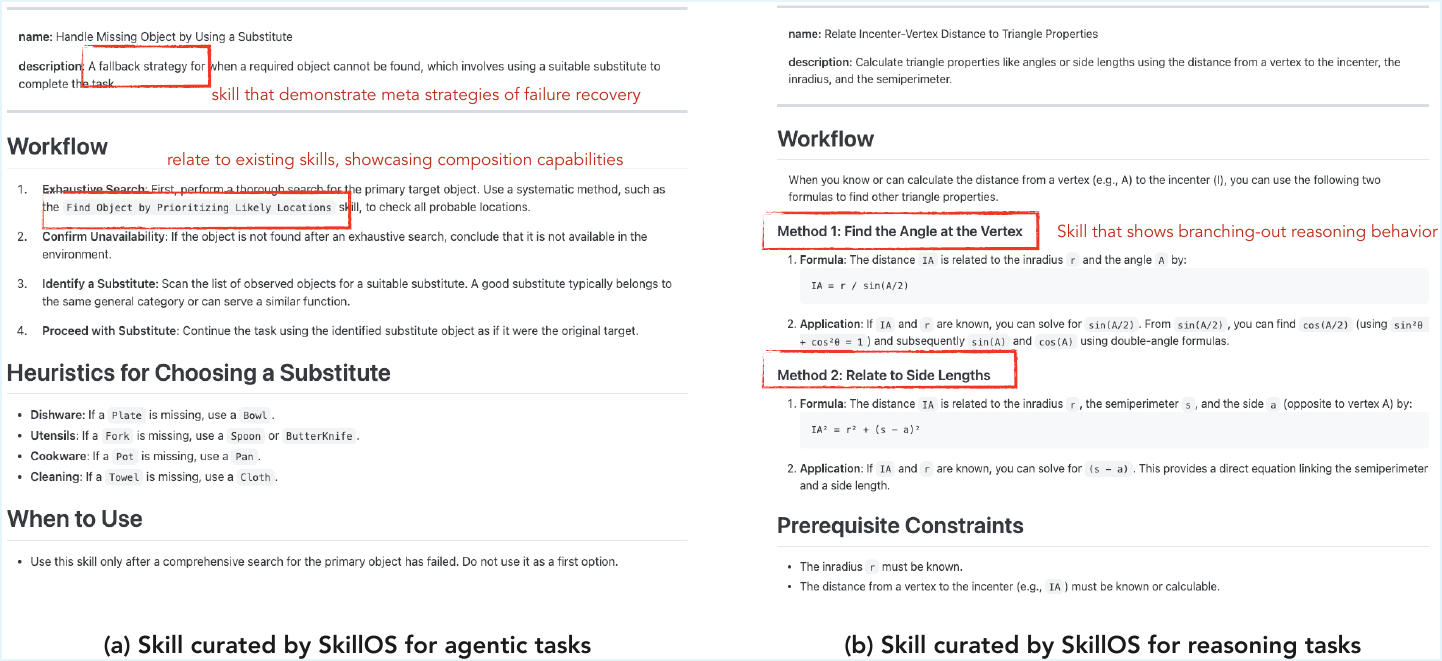}
\end{center}

\caption{Case studies of curated skills by \ours{}.}
\label{fig: skill_case}
\vspace{-3mm}
\end{figure}

\paragraph{Curated Skills for Different Tasks.} Figure~\ref{fig: skill_case} presents two representative skills curated by \ours{} that illustrate qualitatively different curation patterns across task types.
For agentic tasks (Figure~\ref{fig: skill_case}(a)), the curator distills a meta-strategy for failure recovery: rather than memorizing a specific object-search trajectory, it abstracts the recovery procedure into a reusable workflow (\emph{exhaustive search} $\rightarrow$ \emph{confirm unavailability} $\rightarrow$ \emph{identify a substitute} $\rightarrow$ \emph{proceed with substitute}) and explicitly references existing skills, demonstrating compositional curation.
For reasoning tasks (Figure~\ref{fig: skill_case}(b)), the curator captures \emph{branching-out reasoning}: a single skill on inradius--circumradius--semiperimeter relations encodes multiple solution paths (relating the target distance to either the in/circumradius or the side lengths), each paired with its formula, application, and prerequisite constraints.
Together, these examples show that \ours{} learns to produce skills tailored to the structure of the underlying task: procedural and composable for agentic settings, and multi-path with explicit preconditions for reasoning settings, rather than verbatim trajectory copies.

\begin{figure}[t]
\begin{center}
\includegraphics[width=\textwidth]{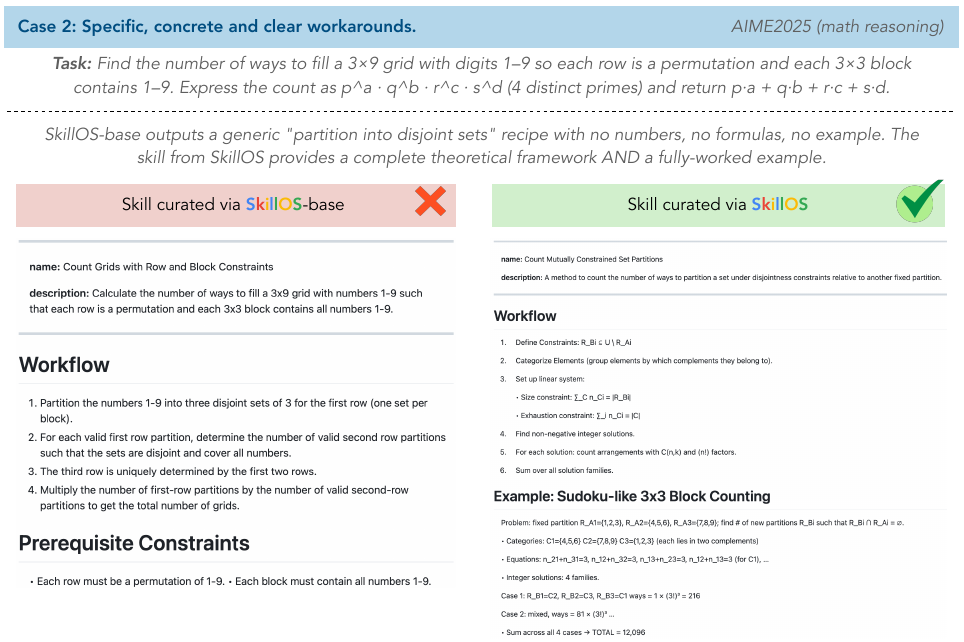}
\end{center}

\caption{Case study on math-reasoning skill curation.
\ours{}-base produces a generic partitioning recipe, while \ours{} curates a concrete and reusable counting framework with explicit constraints, equations, and a worked example.}
\label{fig: case_2}
\vspace{-3mm}
\end{figure}

\paragraph{How \ours{} Curates Better Skills Compared to Baselines.}
We further qualitatively compare the skills curated by \ours{} against those produced by the baseline curator. In the math-reasoning case as shown in Figure~\ref{fig: case_2}, \ours{}-base outputs only a generic high-level recipe based on partitioning into disjoint sets, without explicit formulas, constraints, or examples. By comparison, \ours{} curates a much more useful skill that provides a concrete counting framework, including explicit constraint formulation, equation setup, and a worked example tailored to the target sub-problem. These examples show that RL-trained skill curation improves not only the correctness of the curated content, but also its specificity and usability, enabling skills to better capture the underlying structure of tasks.

\begin{figure}[t]
\begin{center}
\includegraphics[width=\textwidth]{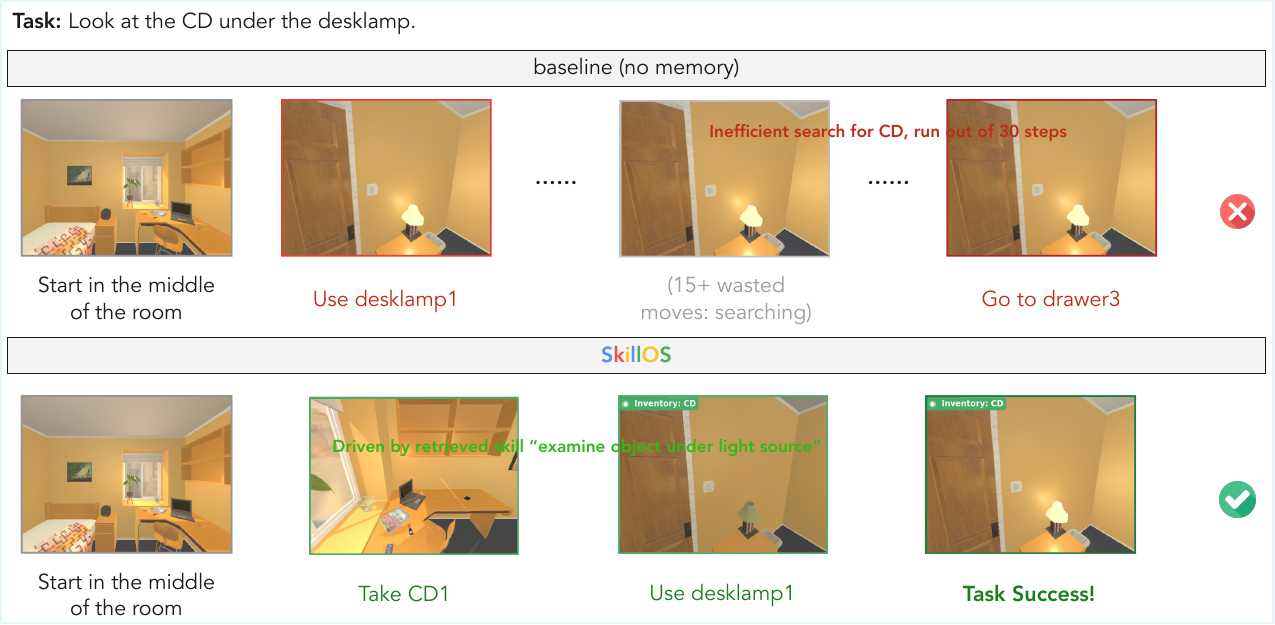}
\end{center}

\caption{Case studies of how skills curated by \ours{} successfully helped to solve a task in ALFWorld.}
\label{fig: skill_case_study}
\vspace{-3mm}
\end{figure}

\paragraph{How Curated Skills Help to Solve Tasks Successfully.}
Figure~\ref{fig: skill_case_study} illustrates a representative example of how curated skills improve agent behavior in interactive environments. Given the task ``look at the CD under the desklamp,'' the memory-free baseline fails to infer the correct object--location relation and performs an inefficient search over irrelevant containers, eventually exhausting the step budget. In contrast, \ours{} retrieves a skill that encourages the agent to examine objects under or around light sources when the instruction refers to an object being ``under'' a lamp. Guided by this reusable strategy, the agent first locates and picks up the CD near the desk area, then moves to the desklamp and inspects the correct target location, completing the task successfully. This case highlights that curated skills do not merely memorize task-specific action sequences; instead, they provide transferable decision guidance that helps the agent focus exploration on semantically relevant objects and locations, reducing unnecessary interactions and improving task success.

\section{Limitations}\label{app: limitations}

\paragraph{Retrieval Mechanism.}
Our current implementation relies on a relatively simple keyword-based retrieval mechanism, such as BM25, to retrieve relevant skills from the skill repository. This design choice allows us to isolate the main focus of this work: studying how skills can be curated, updated, and organized through experience-driven learning. However, more advanced retrieval methods, such as dense retrieval, hybrid retrieval, or learned retrievers, may further improve the relevance of retrieved skills and thus lead to stronger downstream performance. We leave the joint optimization of skill curation and skill retrieval to future work.

\paragraph{Simplified Skill Representation.} Following Anthropic's skill paradigm~\citep{anthropic_skills_2025}, we instantiate each skill as a single Markdown file that combines a YAML frontmatter and Markdown body. This simplification keeps the curator's action space tractable, but it discards two affordances of the original SKILL.md format: (i) supporting scripts and external resource files that allow skills to encapsulate executable procedures rather than purely declarative knowledge, and (ii) hierarchical organization in which a top-level skill can reference or compose lower-level sub-skills. As a result, behaviors that are most naturally expressed as runnable code or as compositions of finer-grained primitives must currently be flattened into prose. Extending \ours{} to multi-file, hierarchical, and partially executable skills is a natural next step.

\paragraph{Frozen Agent Executor.} Throughout training, we keep the agent executor $\pi_{\mathcal{L}}$ frozen and optimize only the skill curator $\pi_{\mathcal{S}}$. This decoupling is deliberate: it isolates the contribution of skill curation, makes the recipe modular across executors, and avoids confounding our analysis with executor-side adaptation. The downside is that the curator can only shape the system's behavior through what it writes into \textsc{SkillRepo}; any miscalibration between the curated skills and the executor's idiosyncrasies must be absorbed by the curator alone. Joint or alternating optimization of $\pi_{\mathcal{S}}$ and $\pi_{\mathcal{L}}$ may yield a better-aligned pair, at the cost of executor specificity and substantially higher training cost.

\section{Future Research Directions}

Our work opens several promising directions for future research.

\paragraph{Agentic Search over Experiential Memory.}
\ours{} currently retrieves relevant skills from \textsc{SkillRepo} through a fixed top-$k$ BM25 lookup, treating retrieval as a static, one-shot operation. As the skill repository grows across thousands of tasks and domains, the bottleneck of self-evolving agents shifts from \textit{what to store} to \textit{how to reliably retrieve and inject the right fragments} at each decision step. A natural next step is to replace static retrieval with \textbf{agentic search}: letting the Skill Curator (or a dedicated retrieval agent) actively issue multiple queries, reformulate them based on intermediate evidence, and iteratively decide which skills to surface, cite, or compose for the executor. This reframes memory access as a first-class decision in the agent's policy rather than a preprocessing step, and opens the door to scaling \ours{} to memory stores orders of magnitude larger than those considered here.

\paragraph{Hierarchical and Compositional Skills.}
Our current skills are flat Markdown entries, each describing a single reusable pattern. Real agent competence, however, is hierarchical: high-level procedures invoke lower-level sub-skills, which in turn depend on primitive operations. Extending \textsc{SkillRepo} to support \textbf{hierarchical decomposition} --- where the curator learns not only to insert, update, and delete skills but also to link, compose, and abstract them --- could enable the agent to build increasingly expressive procedural libraries over time. This direction connects naturally to program-synthesis and library-learning literature, and would allow \ours{} to scale to longer-horizon tasks where single-skill retrieval is insufficient.

\paragraph{Multi-Agent and Shared Memory.}
\ours{} treats memory as a single agent's private artifact. In many realistic deployments, however, multiple agents operate in parallel (e.g., code review, multi-hop research, collaborative robotics) and could benefit from \textbf{shared experiential memory}. Open questions include how to arbitrate conflicting curation decisions from different agents, how to attribute credit when a shared skill contributes to one agent's success but another's failure, and how to preserve specialization while enabling cross-agent transfer. Our GRPO-based curator provides a natural starting point, but extending it to the multi-agent credit-assignment setting is non-trivial and likely to require new algorithmic ideas.

\section{Use of LLMs}
We used LLMs as a general-purpose writing assist tool during the preparation of this submission. Specifically, LLMs were employed for polishing the clarity and readability of text (e.g., refining sentence structure, improving grammar, and shortening overly verbose phrasing). All research ideas, methodology design, experiments, analyses, and final writing decisions were conceived, implemented, and validated solely by the authors.

\end{document}